\documentclass{llncs}

\usepackage{graphicx}
\usepackage{amsmath}
\usepackage{amssymb}
\usepackage{subcaption}
\usepackage{booktabs}
\usepackage{fancybox}
\usepackage{xcolor}

\usepackage[numbers]{natbib}
\begin{document}
%
%
\pagestyle{headings}  
%
\title{Distribution Matching Losses Can Hallucinate Features in Medical Image Translation}
\titlerunning{Distribution Matching Losses Can Hallucinate Features in Medical Image Translation}  

\authorrunning{Cohen et al.} 
%
\tocauthor{Joseph Paul Cohen, Margaux Luck, Sina Honari}
\author{Joseph Paul Cohen, Margaux Luck, Sina Honari}

\institute{Montreal Institute for Learning Algorithms, University of Montreal\\
 \{cohenjos, luckmarg, honaris\}@iro.umontreal.ca}

\maketitle              

\begin{abstract}
This paper discusses how distribution matching losses, such as those used in CycleGAN, when used to synthesize medical images can lead to mis-diagnosis of medical conditions. It seems appealing to use these new image synthesis methods for translating images from a source to a target domain because they can produce high quality images and some even do not require paired data. However, the basis of how these image translation models work is through matching the translation output to the distribution of the target domain. 
This can cause an issue when the data provided in the target domain has an over or under representation of some classes (e.g. healthy or sick).
When the output of an algorithm is a transformed image there are uncertainties whether all known and unknown class labels have been preserved or changed. Therefore, we recommend that these translated images should not be used for direct interpretation (e.g. by doctors) because they may lead to misdiagnosis of patients based on hallucinated image features by an algorithm that matches a distribution. However there are many recent papers that seem as though this is the goal.
\end{abstract}

\section{Introduction}
The introduction of adversarial losses \cite{Goodfellow2014} made it possible to train new kinds of models based on implicit distribution matching. Recently, adversarial approaches such as CycleGAN \cite{ZhuCycleGAN2017}, 
pix2pix \cite{Isola2017}, UNIT \cite{Liu2017}, Adversarially Learned Inference (ALI) \cite{Dumoulin}, and GibbsNet \cite{Lamb2017} have been proposed for un-paired and paired image translation between two domains.
These approaches have been used recently in medical imaging research for translating images between domains such as MRI and CT. However, there is a bias when the output of these models are used for interpretation. When translating images from a source domain to a target domain, these models are trained to match the target domain distribution, where they may hallucinate images by adding or removing image features. 
This can cause a problem when the target distribution during training has over or under representation of known or unknown labels compared to the test time distribution. Due to such a bias, we recommend until better solutions are proposed that maintain the vital information, such translated images should not be used for medical diagnosis, since they can lead to mis-diagnosis of medical conditions.
This issue should be discussed because recently several papers have been published performing image translation using distribution matching.
The main motivation for many of these approaches was to translate images from a source domain to a target domain such that they could be later used for interpretation (e.g. by doctors).
Applications include MR to CT \cite{Wolterink2017a, Nie2016}, CS-MRI \cite{Quan2018, Yang2018}, CT to PET \cite{Ben-Cohen2017}, and  automatic H\&E staining \cite{Bayramolu2017}.

\begin{figure}[t]
    \centering
    \begin{subfigure}[b]{0.5\textwidth}
        \centering
        \includegraphics[width=1\textwidth]{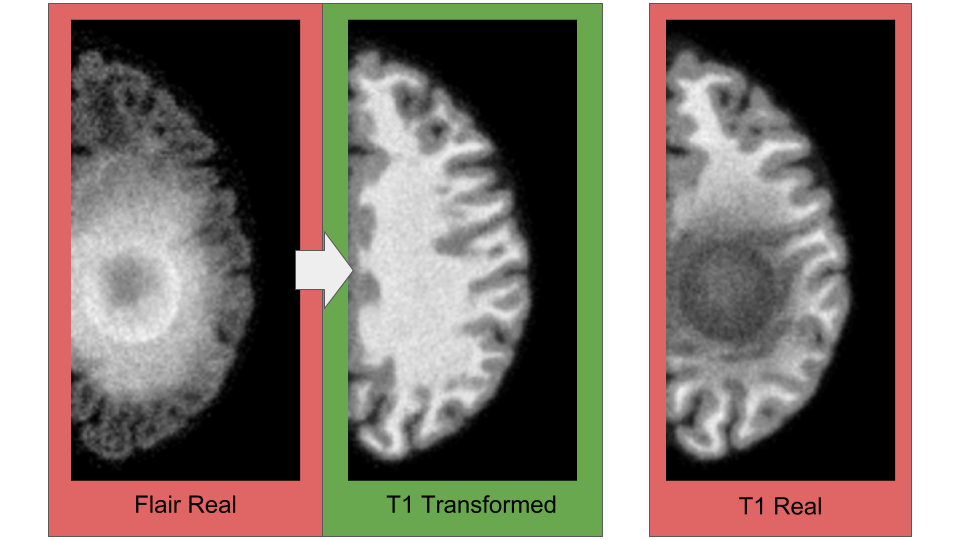}
        \caption{A translation removing tumors}
    \end{subfigure}%
    \begin{subfigure}[b]{0.5\textwidth}
    \centering
        \includegraphics[width=1\textwidth]{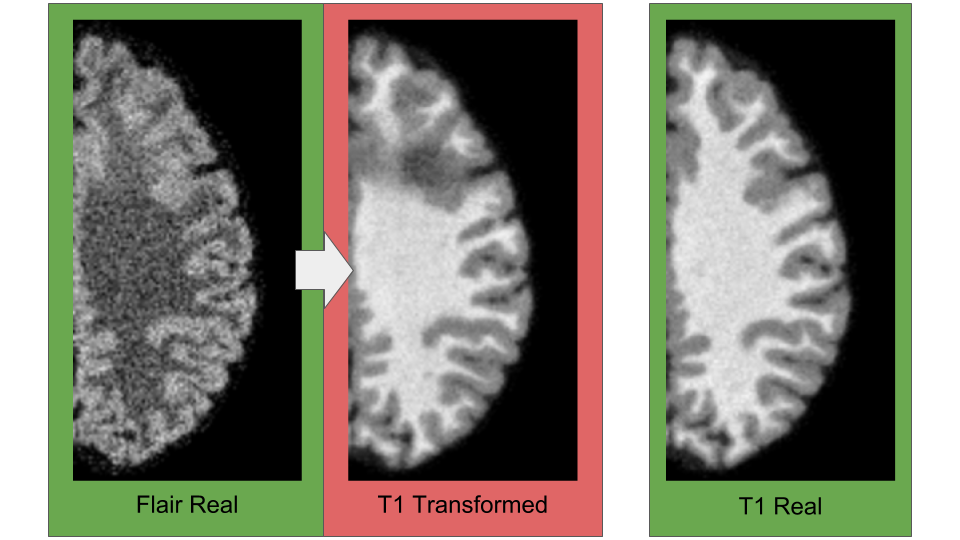}
        \caption{A translation adding tumors}
    \end{subfigure}
    \definecolor{ggreen}{rgb}{0.1, 0.5, 0.0}
    \vspace{-5pt}
    \caption{\small Examples of two CycleGANs trained to transform MRI images from Flair to T1 types. We show healthy images in {\color{ggreen}green} and tumor images in {\color{red}red}. In (a) the model was trained with a bias to remove tumors because the target distribution did not have any tumor examples so the transformation was forced to remove tumors in order to match the target distribution. Conversely in (b) the tumors were added to the image to match the distribution which was composed of only tumor examples during training.}
    \label{fig:removetumor}
    \vspace{-10pt}
\end{figure}

We demonstrate the problem with a caricature example in Figure \ref{fig:removetumor} where we \textit{cure cancer} (in images) and \textit{cause cancer} (in images) using a CycleGAN that translates between Flair and T1 MRI samples. In Figure  \ref{fig:removetumor}(a) the model has been trained only on healthy T1 samples which causes it to remove cancer from the image. This model has learned to match the target distribution regardless of maintaining features that are present in the image. In the following sections, we demonstrate how these methods introduce a bias in image translation due to matching the target distribution. 

We draw attention to this issue in the specific use case where the images are presented for interpretation. However, we do not aim to discourage work using these losses for data augmentation to improve the performance of a classification, segmentation, or other model.

\section{Problem Statement}

Our argument is that the composition of the source and target domains can bias the image transformation to cause an unwanted feature hallucination. We systematically review the objective functions used for image translation in Table \ref{tab:loss} and discuss how they each exhibit this bias.

Let's first consider a standard GAN model \cite{Goodfellow2014} where the generator is a transformation function $f_{a,b}(a)$ which maps samples from the source domain $D_a$ to samples from the target domain $D_b$.  The discriminator is trained given samples from $D_b$ through which the transformation function can match the distribution of $D_b$. 

$$\text{GAN Disc: }\max \underset{b \sim {\color{red}D_b}}{\mathbb{E}} [log D(b)] + 
\underset{a \sim D_a}{\mathbb{E}} [log(1 - D(f_{a,b}(a)))]$$
\noindent In order to minimize this objective the transformation function will need to produce images that match real images from the distribution $D_b$. Here there are no constraints to force a correct mapping between $D_a$ and $D_b$, so for a non-finite $D_a$ we can consider it to be equal to a Gaussian noise $\mathcal{N}$ typically used in a GAN. 

In order to better enforce the mapping between the domains CycleGAN \cite{ZhuCycleGAN2017} extends the generator loss to include cycle consistency terms:
$$ \text{Cycle Consistency: } \left|f_{b,a}(f_{a,b}(a)) - a \right|$$
\noindent Here the function $f_{a,b}$ is composed of the inverse transformation $f_{b,a}$ to create a reconstruction loss that will regularize both transformations to not ignore the source image. However, this process does not provide a guarantee that a correct mapping will be made. In order to match the target distribution, image features can be hallucinated and information to reconstruct an image in the other domain can be encoded \cite{Chu2017}. Moreover, due to having un-paired source and target data, the target distribution that the generator is trained on may be even distinct from the target distribution that corresponds to the data in the source domain (e.g. having only tumor targets while the source is all healthy). This makes the models such as CycleGAN even more prone to hallucinate features due to the way the data in the target domain is gathered. 

Another approach to solve this problem is using a conditional discriminator \cite{Isola2017,mirza2014conditional}. The intuition here is that giving the discriminator the source image $a$ as well as the transformed image $f_{a,b}(a)$, we can model the joint distribution. This approach requires paired examples in order to provide real source and target pairs to the discriminator. The dataset $D_b$ still plays a role in determining what the discriminator learns and therefore how the transformation function operates. The discriminator is trained by:
$$\max \underset{(a,b) \sim (D_a, {\color{red}D_b})}{\mathbb{E}} [log D(b,a)] + 
\underset{a \sim D_a}{\mathbb{E}} [log(1 - D(f_{a,b}(a),a))] $$
Even in the case of CondGAN that the source and target domain distributions correspond to each other due to having paired data, the discriminator can assign more/less capacity to a feature (e.g. tumors), due to having over/under representation of those features in the target distribution. This can be a source of bias in how those features are translated. 

Finally, we look at how to train a transformation using only a L1 loss without any adversarial distribution matching term. With this classic approach we consider transformations based on minimizing the pixel wise error:
$$\underset{(a,b) \sim (D_a, {\color{red}D_b})}{\mathbb{E}} || f_{a,b}(a) - b ||_1$$
Unlike GAN models that match the target distribution over the entire image, L1 predicts each pixel locally given its receptive field without the need to account for global consistency. As long as some pixels present the category of interest in the image (e.g. tumor), L1 can learn a mapping. However, L1 still can suffer from a bias when the train and test distributions are different, e.g. when no tumor pixels are provided during training, which can be caused by having new known or unknown labels at test time.

With all these approaches to domain translation we find there is the potential for bias in the training data (specifically $D_b$ for our experiments below).

\begin{table}[t]
\caption{Loss formulations divided into two phases of training. On the left the discriminator loss is shown (when applicable) and on the right the transformation/generator loss is shown. Note that for GAN losses the generator matches the target distribution indirectly through gradients it receives from the discriminator.} 
\label{tab:loss}

\centering
\resizebox{1\linewidth}{!}{
\begin{tabular}{c c c}
\toprule
& Discriminator Loss (max) & Domain Transformer/Generator Loss (min) \\
\midrule
GAN & 
$\underset{b \sim {\color{red}D_b}}{\mathbb{E}} [log D(b)] + 
\underset{a \sim D_a}{\mathbb{E}} [log(1 - D(f_{a,b}(a)))] $ &  
$\underset{a \sim D_a}{\mathbb{E}} [-log(D(f_{a,b}(a)))] $\\
\midrule
CycleGAN & 
$\underset{b \sim {\color{red}D_b}}{\mathbb{E}} [log D(b)] + 
\underset{a \sim D_a}{\mathbb{E}} [log(1 - D(f_{a,b}(a)))] $ & 
$\underset{a \sim D_a}{\mathbb{E}} [-log(D(f_{a,b}(a))) + 
| f_{b,a}(f_{a,b}(a)) - a |] $\\
\midrule
CondGAN & 
$\underset{(a,b) \sim (D_a, {\color{red}D_b})}{\mathbb{E}} [log D(b,a)] + 
\underset{a \sim D_a}{\mathbb{E}} [log(1 - D(f_{a,b}(a),a))] $ & 
$\underset{a \sim D_a}{\mathbb{E}} [-log(D(f_{a,b}(a), a))] $\\
\midrule
L1 & - & $\underset{(a,b) \sim (D_a, {\color{red}D_b})}{\mathbb{E}} || f_{a,b}(a) - b ||_1$ \\
\vspace{-30pt}
\end{tabular}
}
\end{table}

\section{Bias Impact}
\label{sec:bias}
\vspace{-3pt}

We use the BRATS2013 \cite{Menze_et_al_2015} synthetic MRI dataset because we can visually inspect the presence of a tumor, it is freely available to the public, and we have paired data to inspect results. 
Our task for analysis is to transform Flair MRI images (source domain) into T1-weighted images (target domain). We start with 1700 image slices where 50\% are healthy and 50\% have tumors. We use 1400 to construct training sets for the models and 300 as a holdout test set used to test if the transformation added or removed tumors.

In this section, we construct two training scenarios: unpaired and paired. For the CycleGAN we use an unpaired training scenario which keeps the distribution fixed in the source domain (with 50\% healthy and 50\% tumor samples) and changes the ratio of healthy to cancer samples in the target domain $D_b$ to simulate how the distribution matching works when the target distribution is irrelevant to the source distribution. For the CondGAN and L1 models we use a paired training scenario where both the source and target domains have the same proportion of healthy to tumor examples because they have to be presented as pairs to the model. 

\begin{figure}[ht]
    \centering
    \begin{subfigure}[b]{0.333\textwidth}
        \includegraphics[width=1.0\textwidth]{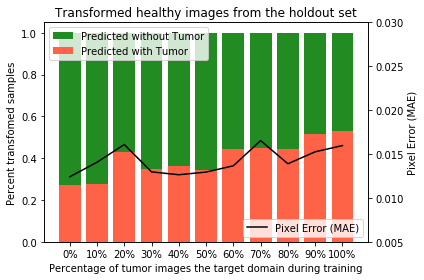}
        \includegraphics[width=1.0\textwidth]{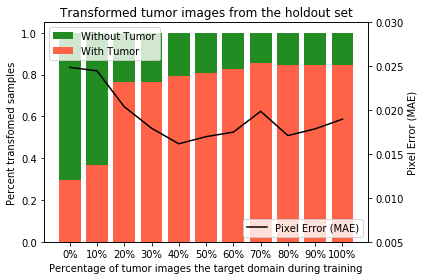}%
        \caption{CycleGAN}
    \end{subfigure}%
    \begin{subfigure}[b]{0.333\textwidth}
        \includegraphics[width=1.0\textwidth]{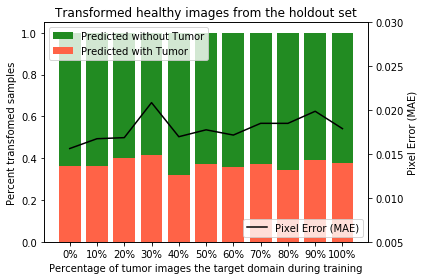}
        \includegraphics[width=1.0\textwidth]{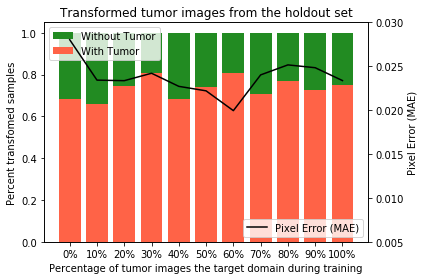}%
        \caption{CondGAN}
    \end{subfigure}%
        \begin{subfigure}[b]{0.333\textwidth}
        \includegraphics[width=1.0\textwidth]{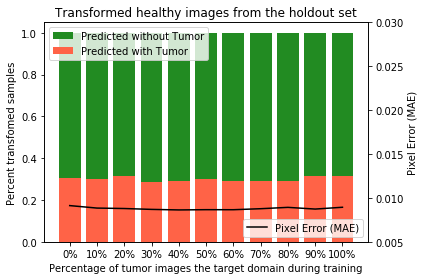}
        \includegraphics[width=1.0\textwidth]{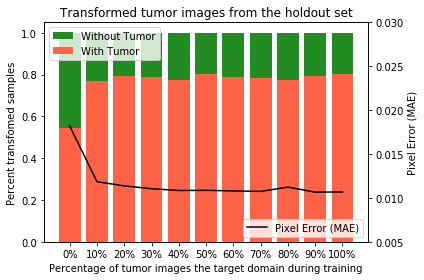}%
        \caption{L1}
    \end{subfigure}%

    \definecolor{ggreen}{rgb}{0.1, 0.5, 0.0}
    \caption{\small We plot the classifier's prediction on 300 (53\% tumor) unseen samples (holdout test set) as we vary the distribution of tumor samples in the target domain from 0\% to 100\% of three models (CycleGAN, CondGAN, L1). This corresponds to 33 trained models. We split the source domain samples of the holdout test set into healthy (top row) and tumor (bottom row) and apply a classifier on the translated images. {\color{ggreen}Green} represents translated samples predicted by the classifier as healthy and {\color{red}red} represents samples predicted with tumors. If the translation was without bias the percentage of healthy to tumor images should not change across the 11 models trained for each loss. For CycleGAN, we observe that the percentage of the images diagnosed with tumors increases as the percentage of tumor images in the target distribution increases. The black line represents the mean absolute pixel error between translated and ground truth target samples. 
    While CondGAN seems to have a more stable classification results compared to CycleGAN, the pixel error indicates how much the translated images are away from ground truth samples and subject to change for different percentage of tumor composition in the target domain. L1 loss seem to suffer the least from target distribution matching and produces high error only when the target distribution has 0\% of tumors (during training) and is asked to translate tumor samples. This case corresponds to 0\% L1 on the bottom row.}
    \label{fig:varydist}
    \vspace{-5pt}
\end{figure}

We train 3 models under 11 different percentages of tumor examples in the target distribution, which vary from 0\% to 100\% with tumors. 
In place of a doctor to classify the transformed samples we use an impartial CNN classifier (4 convolutional layers with ReLU and Stride-2 convolutions, 1 fully connected layer with no non-linearity, and a two-way softmax output layer) which obtains 80\% accuracy on the test set. The results of using this classifier on the generated T1 samples with different target domain composition is shown in Figure \ref{fig:varydist}. 
As we change the composition of the target domain we can observe the bias impact on the class of the transformed examples from the holdout test set.
If there was no bias in matching the target distribution due to the composition of the samples in the target domain, there would be no difference in the percentage of the images diagnosed with a tumor as we change the target domain composition in Figure \ref{fig:varydist}.
We also compute the mean absolute pixel reconstruction error between the ground truth image in the target domain and the translated image. If a large feature is added or removed it should produce a large pixel error. If the translation was doing a perfect job, the pixel error should have been 0 for all cases.

We draw the readers attention to CycleGAN which produces the most dramatic change in class labels, since the model learns to map a balanced (tumor to healthy) source domain to an unbalanced composition in the target domain, which encourages the model to add or remove features (see samples in Figure \ref{fig:varyb}). 
This indicates such models are subject to even more bias due to the composition of the features in the target domain that can be different from the ones in the source domain. 

\begin{figure}
    \captionsetup[subfigure]{labelformat=empty}
    \centering
    
    \begin{subfigure}[b]{1.0\textwidth}

        \caption{(a) An example with a tumor  from the holdout test set}
    \begin{subfigure}[b]{0.03\textwidth}\rotatebox{90}{\hspace{5pt}CycleGAN}\end{subfigure}\hspace{0pt}%
    \begin{subfigure}[b]{0.07\textwidth}\includegraphics[width=\linewidth, trim={0 50 90 50}, 
    clip]{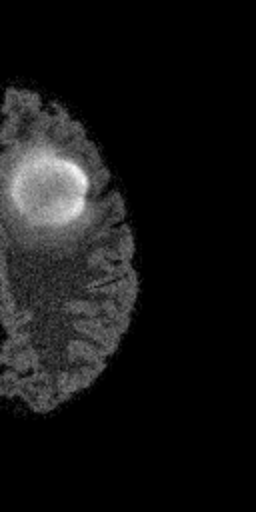}\end{subfigure}\hspace{10pt}%
    \begin{subfigure}[b]{0.07\textwidth}\includegraphics[width=\linewidth, trim={0 50 90 50}, clip]{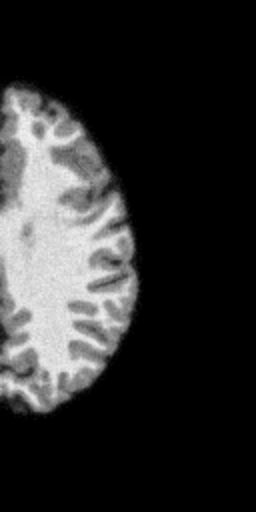}\end{subfigure}%
    \begin{subfigure}[b]{0.07\textwidth}\includegraphics[width=\linewidth, trim={0 50 90 50}, clip]{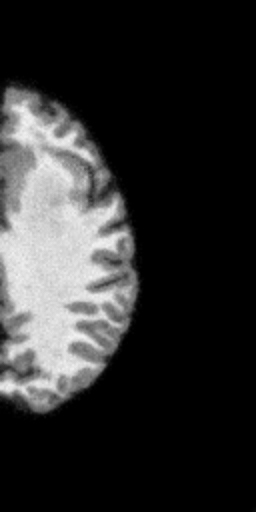}\end{subfigure}%
    \begin{subfigure}[b]{0.07\textwidth}\includegraphics[width=\linewidth, trim={0 50 90 50}, clip]{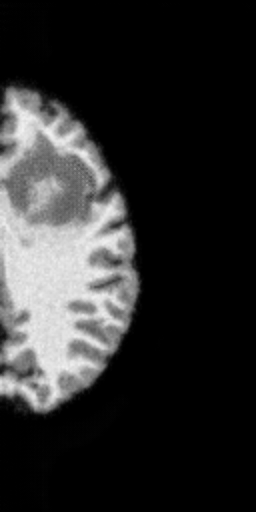}\end{subfigure}%
    \begin{subfigure}[b]{0.07\textwidth}\includegraphics[width=\linewidth, trim={0 50 90 50}, clip]{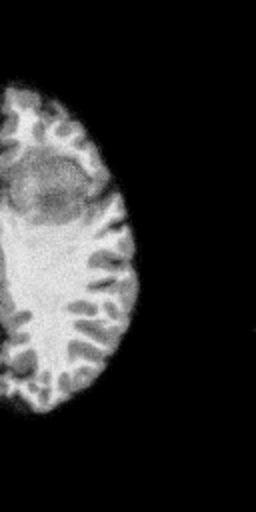}\end{subfigure}%
    \begin{subfigure}[b]{0.07\textwidth}\includegraphics[width=\linewidth, trim={0 50 90 50}, clip]{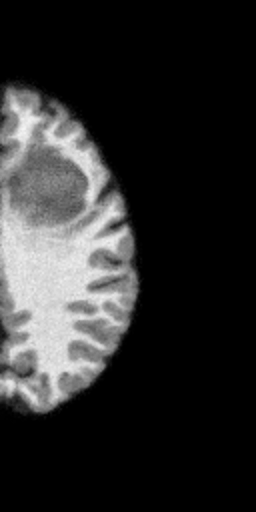}\end{subfigure}%
    \begin{subfigure}[b]{0.07\textwidth}\includegraphics[width=\linewidth, trim={0 50 90 50}, clip]{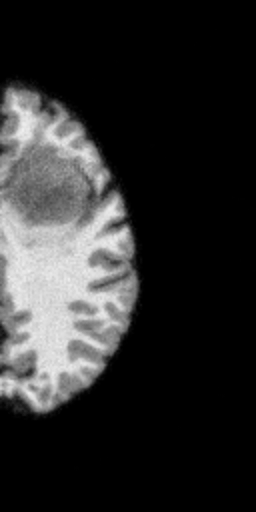}\end{subfigure}%
    \begin{subfigure}[b]{0.07\textwidth}\includegraphics[width=\linewidth, trim={0 50 90 50}, clip]{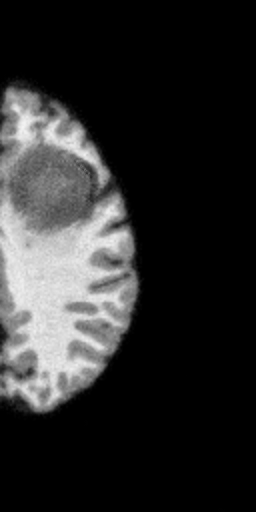}\end{subfigure}%
    \begin{subfigure}[b]{0.07\textwidth}\includegraphics[width=\linewidth, trim={0 50 90 50}, clip]{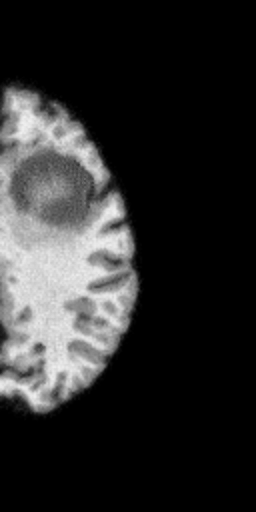}\end{subfigure}%
    \begin{subfigure}[b]{0.07\textwidth}\includegraphics[width=\linewidth, trim={0 50 90 50}, clip]{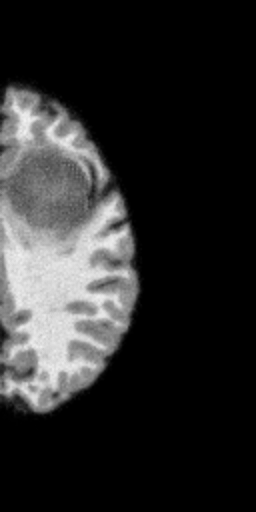}\end{subfigure}%
    \begin{subfigure}[b]{0.07\textwidth}\includegraphics[width=\linewidth, trim={0 50 90 50}, clip]{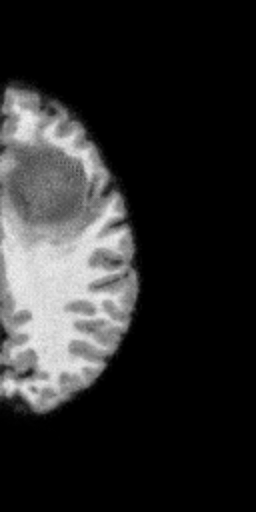}\end{subfigure}%
    \begin{subfigure}[b]{0.07\textwidth}\includegraphics[width=\linewidth, trim={0 50 90 50}, clip]{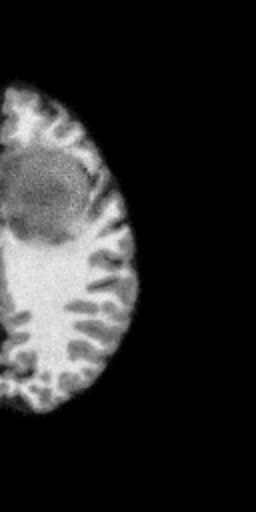}\end{subfigure}%
    \hspace{10pt}\begin{subfigure}[b]{0.07\textwidth}\includegraphics[width=\linewidth, trim={0 50 90 50}, 
    clip]{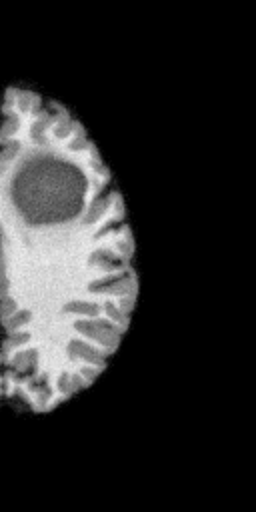}\end{subfigure}%
    
    \begin{subfigure}[b]{0.03\textwidth}\rotatebox{90}{\hspace{4pt}CondGAN}\end{subfigure}\hspace{0pt}%
    \begin{subfigure}[b]{0.07\textwidth}\includegraphics[width=\linewidth, trim={0 50 90 50}, clip]{Figs/HG0001-103-True_real_A.png}\end{subfigure}\hspace{10pt}%
    \begin{subfigure}[b]{0.07\textwidth}\includegraphics[width=\linewidth, trim={0 50 90 50}, clip]{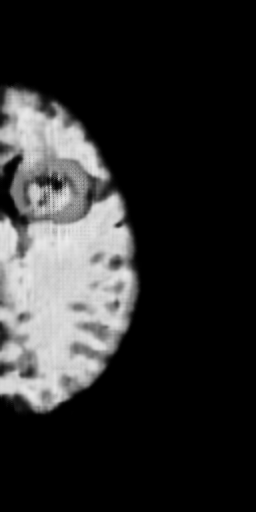}\end{subfigure}%
    \begin{subfigure}[b]{0.07\textwidth}\includegraphics[width=\linewidth, trim={0 50 90 50}, clip]{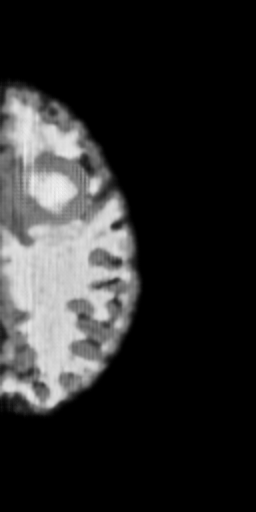}\end{subfigure}%
    \begin{subfigure}[b]{0.07\textwidth}\includegraphics[width=\linewidth, trim={0 50 90 50}, clip]{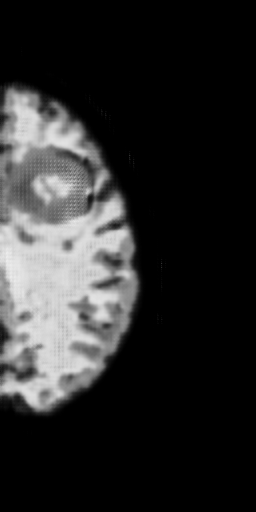}\end{subfigure}%
    \begin{subfigure}[b]{0.07\textwidth}\includegraphics[width=\linewidth, trim={0 50 90 50}, clip]{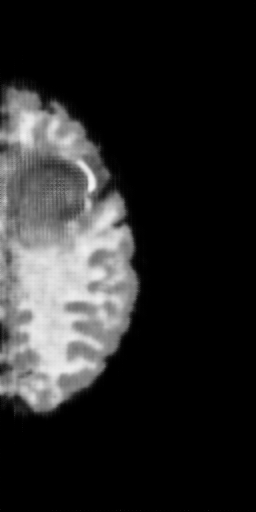}\end{subfigure}%
    \begin{subfigure}[b]{0.07\textwidth}\includegraphics[width=\linewidth, trim={0 50 90 50}, clip]{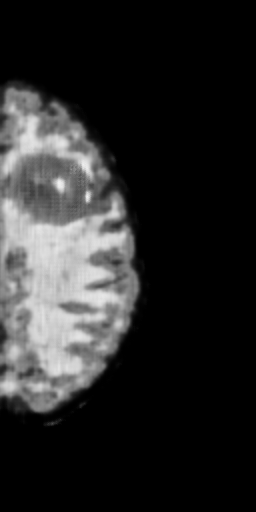}\end{subfigure}%
    \begin{subfigure}[b]{0.07\textwidth}\includegraphics[width=\linewidth, trim={0 50 90 50}, clip]{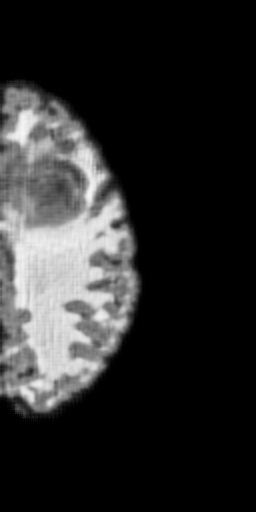}\end{subfigure}%
    \begin{subfigure}[b]{0.07\textwidth}\includegraphics[width=\linewidth, trim={0 50 90 50}, clip]{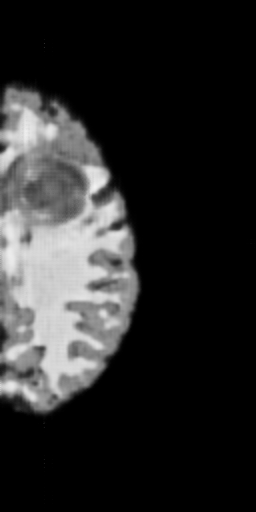}\end{subfigure}%
    \begin{subfigure}[b]{0.07\textwidth}\includegraphics[width=\linewidth, trim={0 50 90 50}, clip]{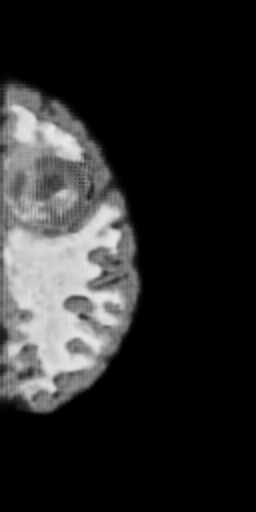}\end{subfigure}%
    \begin{subfigure}[b]{0.07\textwidth}\includegraphics[width=\linewidth, trim={0 50 90 50}, clip]{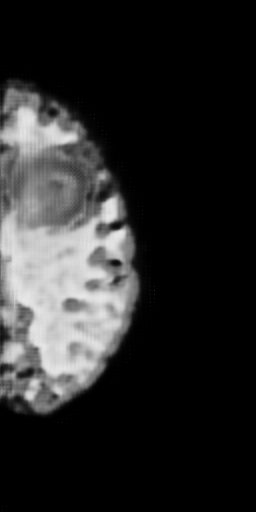}\end{subfigure}%
    \begin{subfigure}[b]{0.07\textwidth}\includegraphics[width=\linewidth, trim={0 50 90 50}, clip]{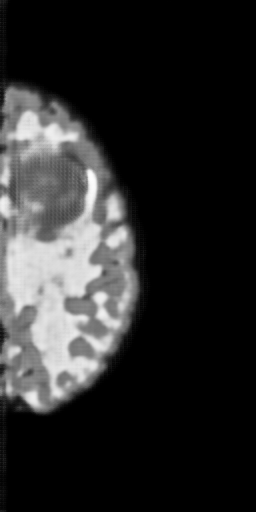}\end{subfigure}%
    \begin{subfigure}[b]{0.07\textwidth}\includegraphics[width=\linewidth, trim={0 50 90 50}, clip]{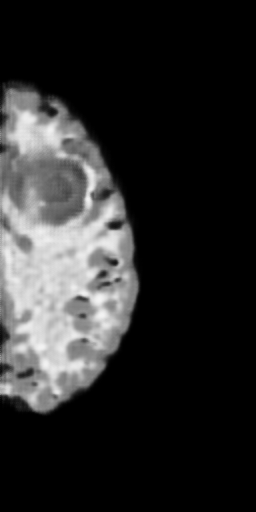}\end{subfigure}%
    \hspace{10pt}\begin{subfigure}[b]{0.07\textwidth}\includegraphics[width=\linewidth, trim={0 50 90 50}, clip]{Figs/HG0001-103-True_real_B.png}\end{subfigure}%
    
    \begin{subfigure}[b]{0.03\textwidth}\rotatebox{90}{\hspace{55pt}L1}\end{subfigure}\hspace{0pt}%
    \begin{subfigure}[b]{0.07\textwidth}\includegraphics[width=\linewidth, trim={0 50 90 50}, clip]{Figs/HG0001-103-True_real_A.png}\caption{\centering \begin{minipage}{1\textwidth}\centering \vspace{2 px} Flair \\ (source) \end{minipage}}\end{subfigure}\hspace{10pt}%
    \begin{subfigure}[b]{0.07\textwidth}\includegraphics[width=\linewidth, trim={0 50 90 50}, clip]{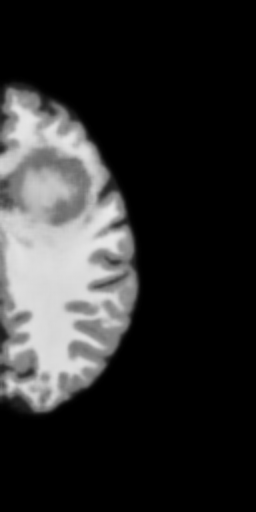}\caption{0\%\\\hspace{1pt}}\end{subfigure}%
    \begin{subfigure}[b]{0.07\textwidth}\includegraphics[width=\linewidth, trim={0 50 90 50}, clip]{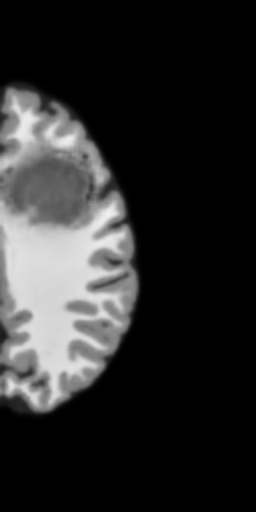}\caption{10\%\\\hspace{1pt}}\end{subfigure}%
    \begin{subfigure}[b]{0.07\textwidth}\includegraphics[width=\linewidth, trim={0 50 90 50}, clip]{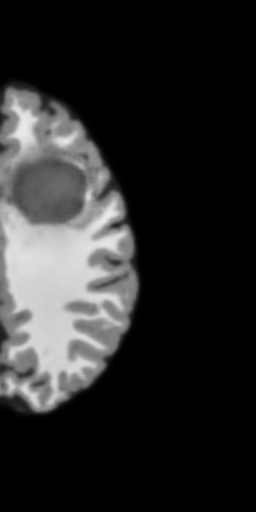}\caption{20\%\\\hspace{1pt}}\end{subfigure}%
    \begin{subfigure}[b]{0.07\textwidth}\includegraphics[width=\linewidth, trim={0 50 90 50}, clip]{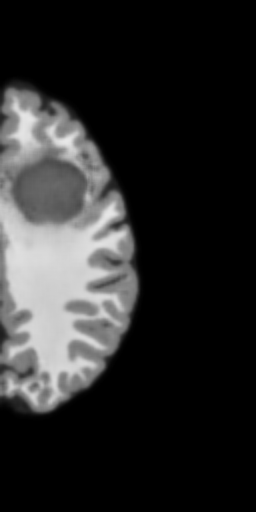}\caption{30\%\\\hspace{1pt}}\end{subfigure}%
    \begin{subfigure}[b]{0.07\textwidth}\includegraphics[width=\linewidth, trim={0 50 90 50}, clip]{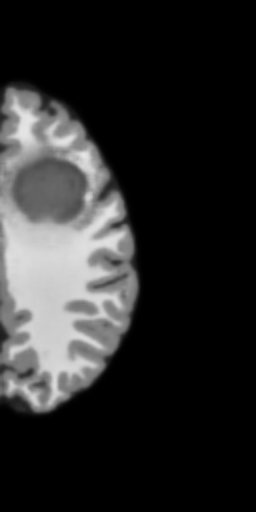}\caption{40\%\\\hspace{1pt}}\end{subfigure}%
    \begin{subfigure}[b]{0.07\textwidth}\includegraphics[width=\linewidth, trim={0 50 90 50}, clip]{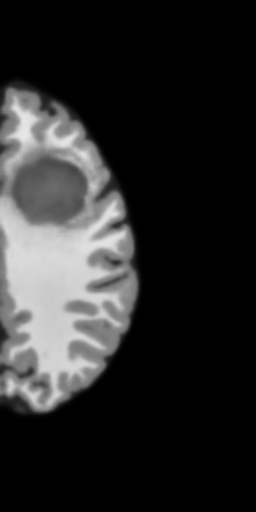}\caption{50\%\\\hspace{1pt}}\end{subfigure}%
    \begin{subfigure}[b]{0.07\textwidth}\includegraphics[width=\linewidth, trim={0 50 90 50}, clip]{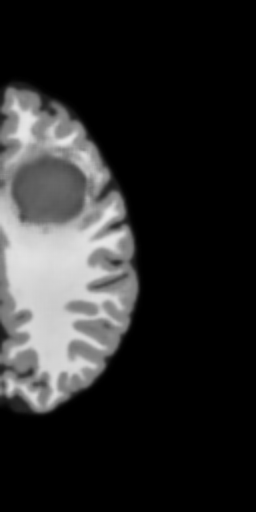}\caption{60\%\\\hspace{1pt}}\end{subfigure}%
    \begin{subfigure}[b]{0.07\textwidth}\includegraphics[width=\linewidth, trim={0 50 90 50}, clip]{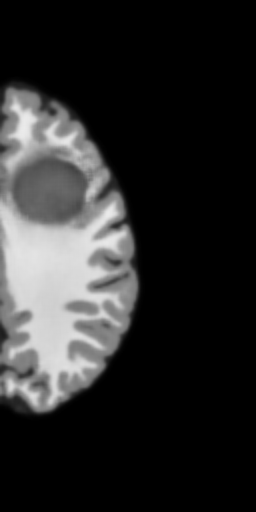}\caption{70\%\\\hspace{1pt}}\end{subfigure}%
    \begin{subfigure}[b]{0.07\textwidth}\includegraphics[width=\linewidth, trim={0 50 90 50}, clip]{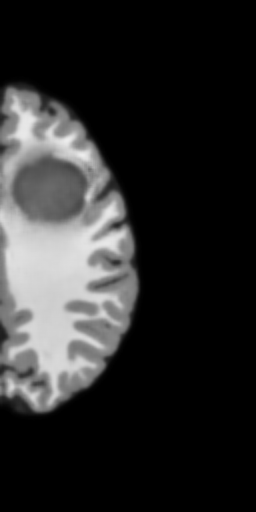}\caption{80\%\\\hspace{1pt}}\end{subfigure}%
    \begin{subfigure}[b]{0.07\textwidth}\includegraphics[width=\linewidth, trim={0 50 90 50}, clip]{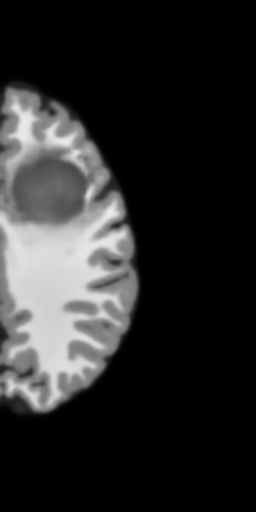}\caption{90\%\\\hspace{1pt}}\end{subfigure}%
    \begin{subfigure}[b]{0.07\textwidth}\includegraphics[width=\linewidth, trim={0 50 90 50}, clip]{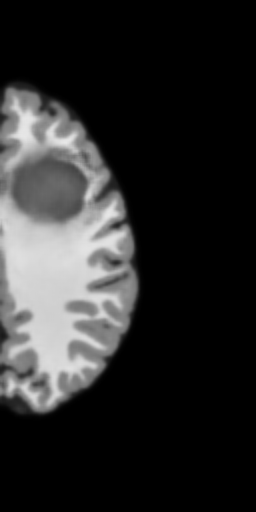}\caption{100\%\\\hspace{1pt}}\end{subfigure}%
    \hspace{10pt}\begin{subfigure}[b]{0.07\textwidth}\includegraphics[width=\linewidth, trim={0 50 90 50}, clip]{Figs/HG0001-103-True_real_B.png}\caption{\centering \begin{minipage}{1\textwidth}\vspace{2 px}\centering{T1\\(target)} \end{minipage}}\end{subfigure}%
    
    \end{subfigure}%

    \begin{subfigure}[b]{1.0\textwidth}
    \caption{(b) A healthy example from the holdout test set}
    \begin{subfigure}[b]{0.03\textwidth}\rotatebox{90}{\hspace{5pt}CycleGAN}\end{subfigure}\hspace{0pt}%
    \begin{subfigure}[b]{0.07\textwidth}\includegraphics[width=\linewidth, trim={0 50 90 50}, 
    clip]{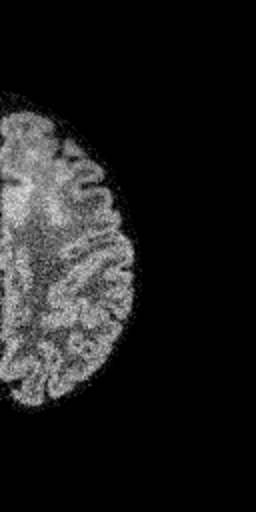}\end{subfigure}\hspace{10pt}%
    \begin{subfigure}[b]{0.07\textwidth}\includegraphics[width=\linewidth, trim={0 50 90 50}, clip]{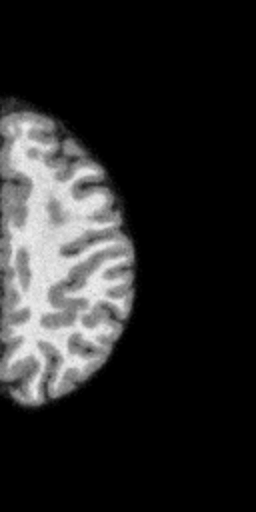}\end{subfigure}%
    \begin{subfigure}[b]{0.07\textwidth}\includegraphics[width=\linewidth, trim={0 50 90 50}, clip]{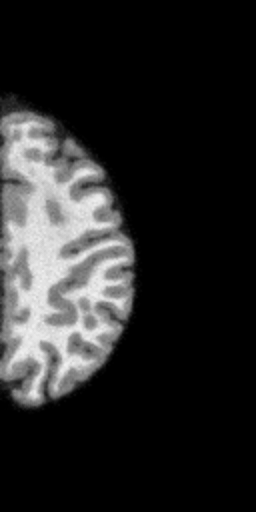}\end{subfigure}%
    \begin{subfigure}[b]{0.07\textwidth}\includegraphics[width=\linewidth, trim={0 50 90 50}, clip]{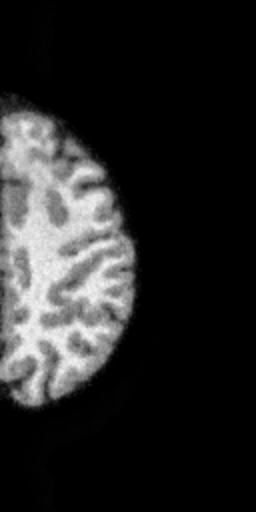}\end{subfigure}%
    \begin{subfigure}[b]{0.07\textwidth}\includegraphics[width=\linewidth, trim={0 50 90 50}, clip]{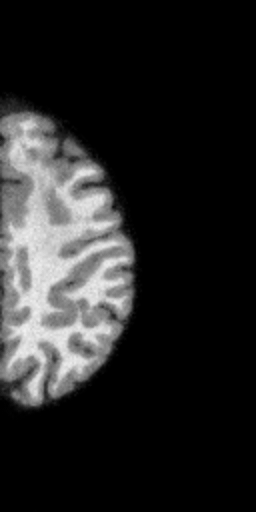}\end{subfigure}%
    \begin{subfigure}[b]{0.07\textwidth}\includegraphics[width=\linewidth, trim={0 50 90 50}, clip]{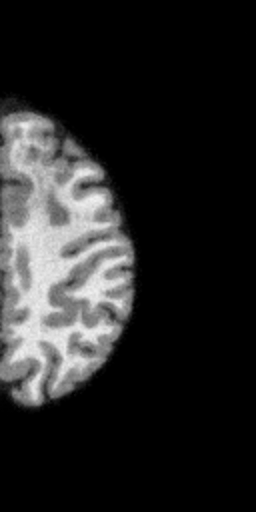}\end{subfigure}%
    \begin{subfigure}[b]{0.07\textwidth}\includegraphics[width=\linewidth, trim={0 50 90 50}, clip]{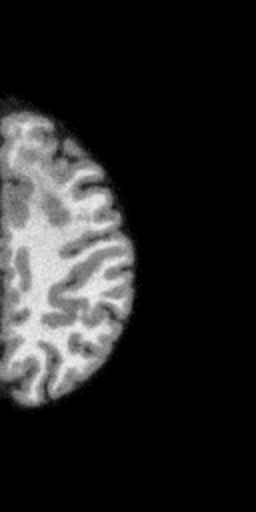}\end{subfigure}%
    \begin{subfigure}[b]{0.07\textwidth}\includegraphics[width=\linewidth, trim={0 50 90 50}, clip]{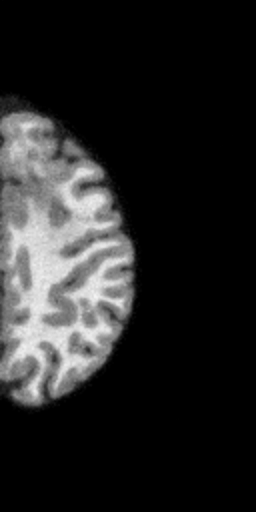}\end{subfigure}%
    \begin{subfigure}[b]{0.07\textwidth}\includegraphics[width=\linewidth, trim={0 50 90 50}, clip]{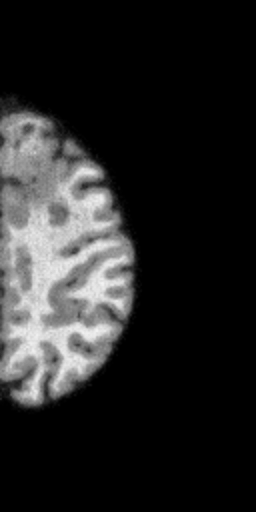}\end{subfigure}%
    \begin{subfigure}[b]{0.07\textwidth}\includegraphics[width=\linewidth, trim={0 50 90 50}, clip]{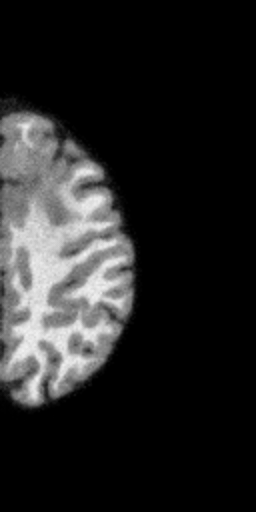}\end{subfigure}%
    \begin{subfigure}[b]{0.07\textwidth}\includegraphics[width=\linewidth, trim={0 50 90 50}, clip]{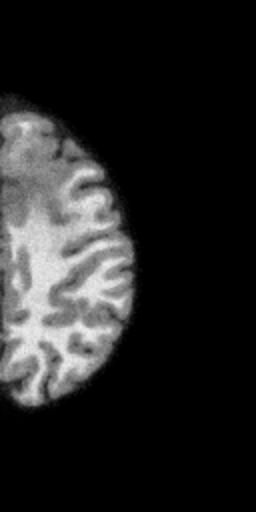}\end{subfigure}%
    \begin{subfigure}[b]{0.07\textwidth}\includegraphics[width=\linewidth, trim={0 50 90 50}, clip]{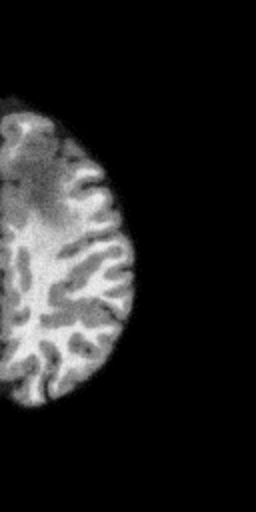}\end{subfigure}%
    \hspace{10pt}\begin{subfigure}[b]{0.07\textwidth}\includegraphics[width=\linewidth, trim={0 50 90 50}, 
    clip]{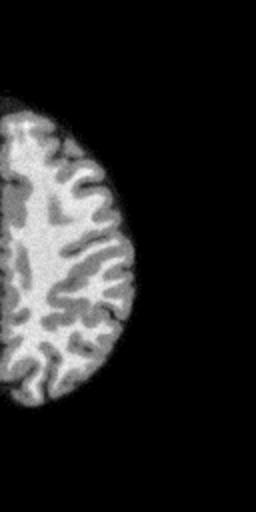}\end{subfigure}%
    
    \begin{subfigure}[b]{0.03\textwidth}\rotatebox{90}{\hspace{4pt}CondGAN}\end{subfigure}\hspace{0pt}%
    \begin{subfigure}[b]{0.07\textwidth}\includegraphics[width=\linewidth, trim={0 50 90 50}, clip]{Figs/LG0012-114-False_real_A.png}\end{subfigure}\hspace{10pt}%
    \begin{subfigure}[b]{0.07\textwidth}\includegraphics[width=\linewidth, trim={0 50 90 50}, clip]{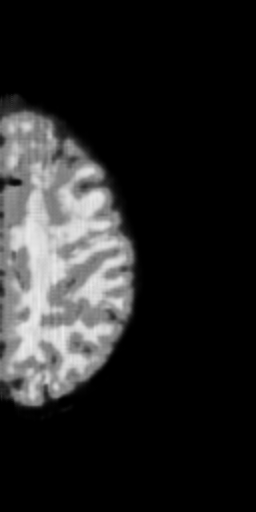}\end{subfigure}%
    \begin{subfigure}[b]{0.07\textwidth}\includegraphics[width=\linewidth, trim={0 50 90 50}, clip]{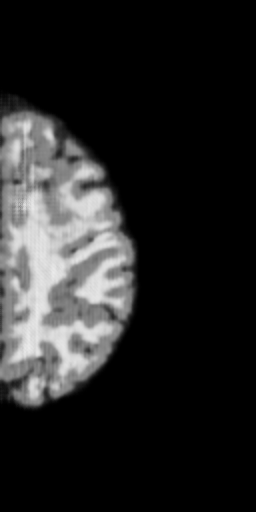}\end{subfigure}%
    \begin{subfigure}[b]{0.07\textwidth}\includegraphics[width=\linewidth, trim={0 50 90 50}, clip]{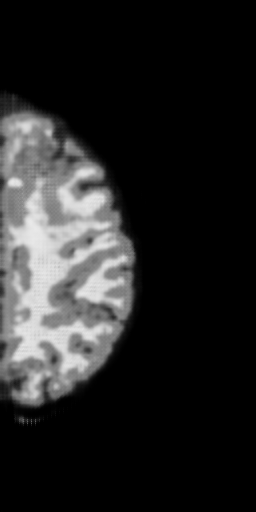}\end{subfigure}%
    \begin{subfigure}[b]{0.07\textwidth}\includegraphics[width=\linewidth, trim={0 50 90 50}, clip]{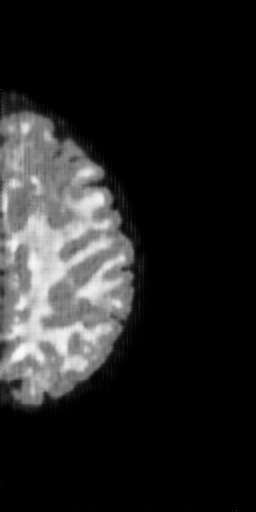}\end{subfigure}%
    \begin{subfigure}[b]{0.07\textwidth}\includegraphics[width=\linewidth, trim={0 50 90 50}, clip]{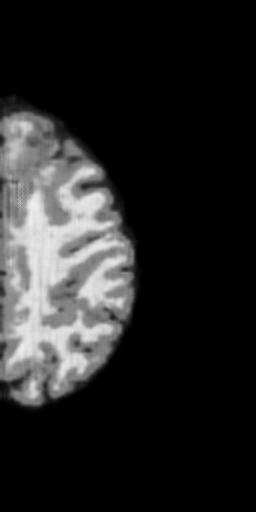}\end{subfigure}%
    \begin{subfigure}[b]{0.07\textwidth}\includegraphics[width=\linewidth, trim={0 50 90 50}, clip]{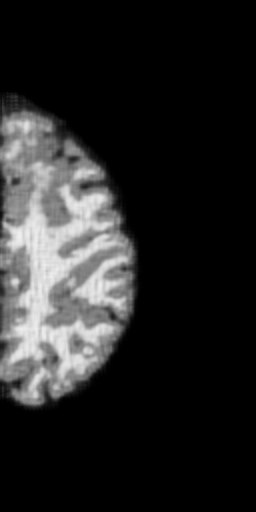}\end{subfigure}%
    \begin{subfigure}[b]{0.07\textwidth}\includegraphics[width=\linewidth, trim={0 50 90 50}, clip]{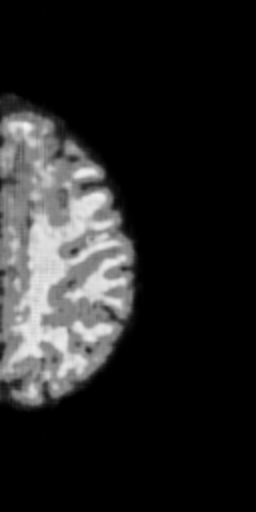}\end{subfigure}%
    \begin{subfigure}[b]{0.07\textwidth}\includegraphics[width=\linewidth, trim={0 50 90 50}, clip]{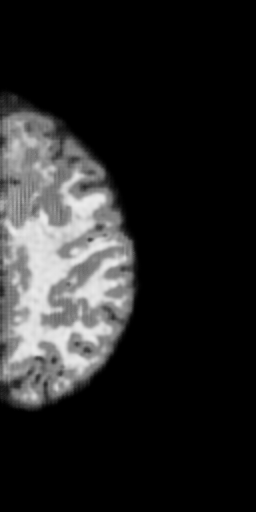}\end{subfigure}%
    \begin{subfigure}[b]{0.07\textwidth}\includegraphics[width=\linewidth, trim={0 50 90 50}, clip]{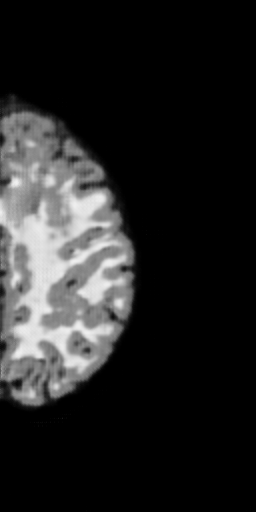}\end{subfigure}%
    \begin{subfigure}[b]{0.07\textwidth}\includegraphics[width=\linewidth, trim={0 50 90 50}, clip]{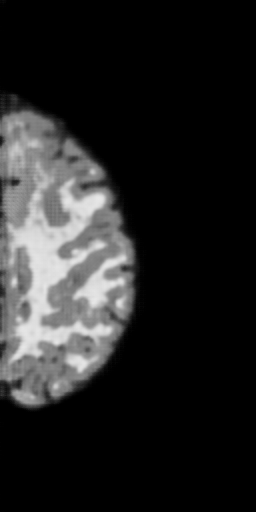}\end{subfigure}%
    \begin{subfigure}[b]{0.07\textwidth}\includegraphics[width=\linewidth, trim={0 50 90 50}, clip]{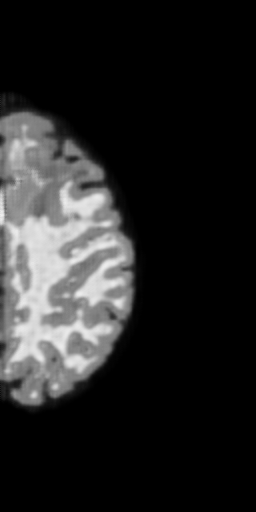}\end{subfigure}%
    \hspace{10pt}\begin{subfigure}[b]{0.07\textwidth}\includegraphics[width=\linewidth, trim={0 50 90 50}, clip]{Figs/LG0012-114-False_real_B.png}\end{subfigure}%
    
    \begin{subfigure}[b]{0.03\textwidth}\rotatebox{90}{\hspace{55pt}L1}\end{subfigure}\hspace{0pt}%
    \begin{subfigure}[b]{0.07\textwidth}\includegraphics[width=\linewidth, trim={0 50 90 50}, clip]{Figs/LG0012-114-False_real_A.png}\caption{\centering \begin{minipage}{1\textwidth}\centering \vspace{2 px} Flair \\ (source) \end{minipage}}\end{subfigure}\hspace{10pt}%
    \begin{subfigure}[b]{0.07\textwidth}\includegraphics[width=\linewidth, trim={0 50 90 50}, clip]{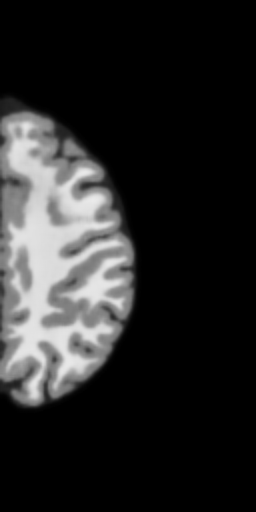}\caption{0\%\\\hspace{1pt}}\end{subfigure}%
    \begin{subfigure}[b]{0.07\textwidth}\includegraphics[width=\linewidth, trim={0 50 90 50}, clip]{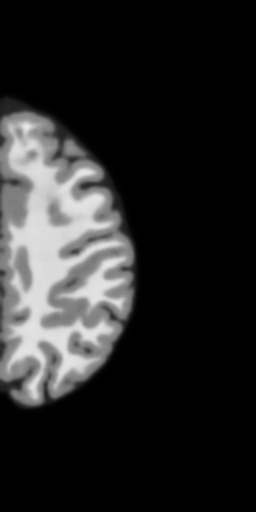}\caption{10\%\\\hspace{1pt}}\end{subfigure}%
    \begin{subfigure}[b]{0.07\textwidth}\includegraphics[width=\linewidth, trim={0 50 90 50}, clip]{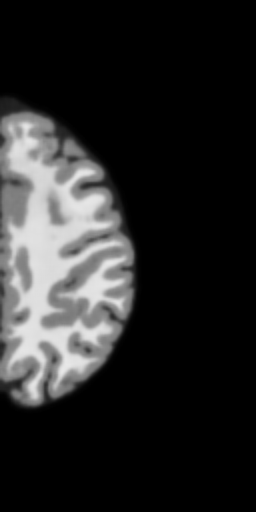}\caption{20\%\\\hspace{1pt}}\end{subfigure}%
    \begin{subfigure}[b]{0.07\textwidth}\includegraphics[width=\linewidth, trim={0 50 90 50}, clip]{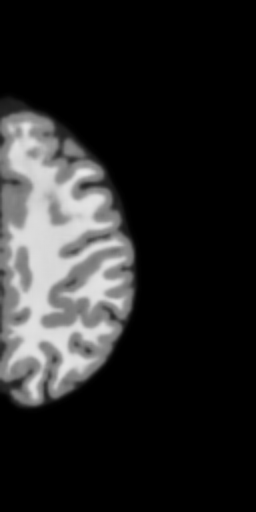}\caption{30\%\\\hspace{1pt}}\end{subfigure}%
    \begin{subfigure}[b]{0.07\textwidth}\includegraphics[width=\linewidth, trim={0 50 90 50}, clip]{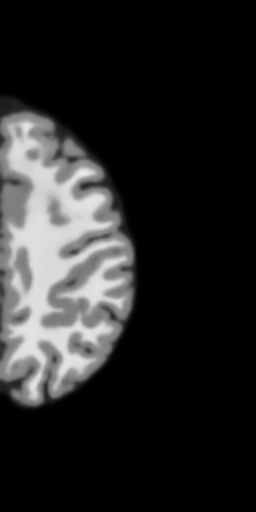}\caption{40\%\\\hspace{1pt}}\end{subfigure}%
    \begin{subfigure}[b]{0.07\textwidth}\includegraphics[width=\linewidth, trim={0 50 90 50}, clip]{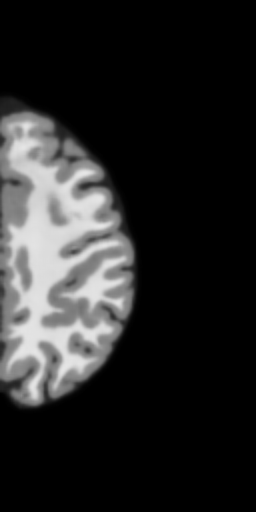}\caption{50\%\\\hspace{1pt}}\end{subfigure}%
    \begin{subfigure}[b]{0.07\textwidth}\includegraphics[width=\linewidth, trim={0 50 90 50}, clip]{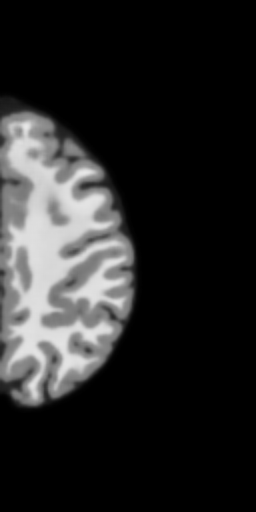}\caption{60\%\\\hspace{1pt}}\end{subfigure}%
    \begin{subfigure}[b]{0.07\textwidth}\includegraphics[width=\linewidth, trim={0 50 90 50}, clip]{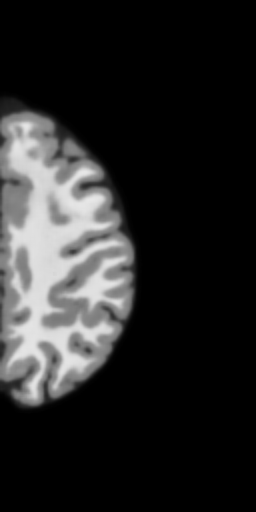}\caption{70\%\\\hspace{1pt}}\end{subfigure}%
    \begin{subfigure}[b]{0.07\textwidth}\includegraphics[width=\linewidth, trim={0 50 90 50}, clip]{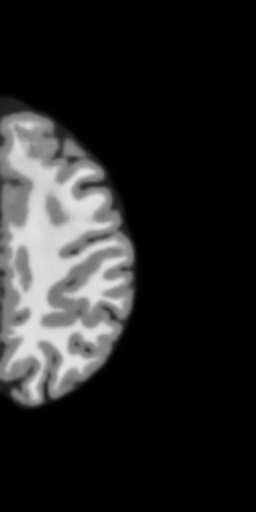}\caption{80\%\\\hspace{1pt}}\end{subfigure}%
    \begin{subfigure}[b]{0.07\textwidth}\includegraphics[width=\linewidth, trim={0 50 90 50}, clip]{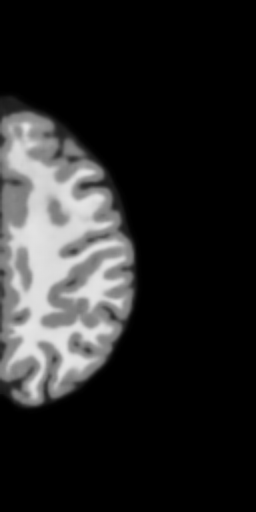}\caption{90\%\\\hspace{1pt}}\end{subfigure}%
    \begin{subfigure}[b]{0.07\textwidth}\includegraphics[width=\linewidth, trim={0 50 90 50}, clip]{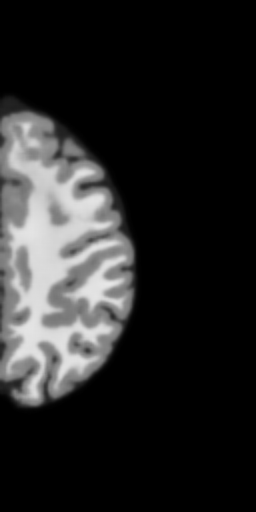}\caption{100\%\\\hspace{1pt}}\end{subfigure}%
    \hspace{10pt}\begin{subfigure}[b]{0.07\textwidth}\includegraphics[width=\linewidth, trim={0 50 90 50}, clip]{Figs/LG0012-114-False_real_B.png}\caption{\centering \begin{minipage}{1\textwidth}\vspace{2 px}\centering{T1\\(target)} \end{minipage}}\end{subfigure}%

    \end{subfigure}%

    \caption{Illustration of tumor (a) and healthy (b) class change through domain translation while changing the ratio of the healthy to tumor samples in the target domain $D_b$ for all three models (CycleGAN, CondGAN, L1). We vary the distribution of $D_b$ from 0\% tumor to 100\% examples to train 33 different models. We show images of the source domain (Flair) on the left and the corresponding ground truth image in the target domain (T1) on the right. We can observe visually the magnitude of the changes introduced.}
    \label{fig:varyb-models}
\end{figure}

For CondGAN, the pixel error changes across as the composition of tumor/healthy changes, indicating there is a bias due to the training data composition.  Perceptually the L1 loss appears the most consistent producing the least bias. However, it has error when it is trained on 0\% tumor and the model is asked to translate tumor samples at test time (0\% for L1 in Figure \ref{fig:varydist} bottom row and Figure \ref{fig:varyb-models} (a)), which is due to a mis-match between train and test distributions. It indicates that if at test time images with new known or unknown labels (e.g. a new disease) are presented to the model, it cannot transform them properly. In Figure \ref{fig:varyb-models} we show examples of the translated images between the models. Note how for GAN based models the cancer tumor gradually appears and gets bigger from left to right. L1 mostly suffers in Figure \ref{fig:varyb-models} (a) for 0\%. Interestingly, in the case of 100\% tumor it can translate healthy images even though it was not trained with healthy images. 
We believe this is due to having both healthy and tumor regions in each image which allows the network to see healthy sub-regions and learn to translate both categories.
Further samples are available in the supplementary information in Figures \ref{fig:bias_class_same} and \ref{fig:bias_class_diff}.

\section{Conclusion}
\vspace{-5pt}

In this work we discussed concerns about how distribution matching losses, such as those used in CycleGAN, can lead to mis-diagnosis of medical conditions.  We have presented experimental evidence that when the output of an algorithm matches a distribution, for unpaired or paired data translation, all known and unknown class labels might not be preserved.
Therefore, these translated images should not be used for interpretation (e.g. by doctors) without proper tools to verify the translation process. We illustrate this problem using dramatic examples of tumors being added and removed from MRI images. We hope that future methods will take steps to ensure that this bias does not influence the outcome of a medical diagnosis.

\section*{Acknowledgements}
We thank Adriana Romero Soriano, Michal Drozdzal, and Mohammad Havaei for their valuable input and assistance on the project.
This work is partially funded by a grant from the U.S.
National Science Foundation Graduate Research Fellowship
Program (grant number: DGE-1356104) and the Institut
de valorisation des donnees (IVADO). This work utilized the supercomputing facilities managed by the Montreal Institute
for Learning Algorithms, NSERC, Compute Canada,
and Calcul Quebec.

{\small
\bibliographystyle{splncs-nopagenum}
\bibliography{gans,medgan,bibliography}

\begin{thebibliography}{10}

\bibitem{Goodfellow2014}
Goodfellow, I.J., Pouget-Abadie, J., Mirza, M., Xu, B., Warde-Farley, D.,
  Ozair, S., Courville, A., Bengio, Y.:
\newblock {Generative Adversarial Networks}.
\newblock  In: Neural Information Processing Systems (2014)

\bibitem{ZhuCycleGAN2017}
Zhu, J.Y., Park, T., Isola, P., Efros, A.A.:
\newblock {Unpaired Image-to-Image Translation using Cycle-Consistent
  Adversarial Networks}.
\newblock  In: International Conference on Computer Vision (2017)

\bibitem{Isola2017}
Isola, P., Zhu, J.Y., Zhou, T., Efros, A.A.:
\newblock {Image-to-image translation with conditional adversarial networks}.
\newblock  In: Computer Vision and Pattern Recognition (2017)

\bibitem{Liu2017}
Liu, M.Y., Breuel, T., Kautz, J.:
\newblock {Unsupervised Image-to-Image Translation Networks}.
\newblock  In: Neural Information Processing Systems (2017)

\bibitem{Dumoulin}
Dumoulin, V., Belghazi, I., Poole, B., Mastropietro, O., Lamb, A., Arjovsky,
  M., Courville, A.:
\newblock {Adversarially Learned Inference}.
\newblock  In: International Conference on Learning Representations (2017)

\bibitem{Lamb2017}
Lamb, A., Hjelm, D., Ganin, Y., Cohen, J.P., Courville, A., Bengio, Y.:
\newblock {GibbsNet: Iterative Adversarial Inference for Deep Graphical
  Models}.
\newblock  In: Neural Information Processing Systems (2017)

\bibitem{Wolterink2017a}
Wolterink, J.M., Dinkla, A.M., Savenije, M.H., Seevinck, P.R., van~den Berg,
  C.A., I{\v{s}}gum, I.:
\newblock {Deep MR to CT synthesis using unpaired data}.
\newblock  In: Workshop on Simulation and Synthesis in Medical Imaging (2017)

\bibitem{Nie2016}
Nie, D., Trullo, R., Petitjean, C., Ruan, S., Shen, D.:
\newblock {Medical Image Synthesis with Context-Aware Generative Adversarial
  Networks}.
\newblock  In: Medical Image Computing and Computer-Assisted Intervention
  (2016)

\bibitem{Quan2018}
Quan, T.M., Nguyen-Duc, T., Jeong, W.K.:
\newblock {Compressed Sensing MRI Reconstruction using a Generative Adversarial
  Network with a Cyclic Loss}.
\newblock IEEE Transactions on Medical Imaging (2018)

\bibitem{Yang2018}
Yang, G., Yu, S., Dong, H., Slabaugh, G., Dragotti, P.L., Ye, X., Liu, F.,
  Arridge, S., Keegan, J., Guo, Y., Firmin, D.:
\newblock {DAGAN: Deep De-Aliasing Generative Adversarial Networks for Fast
  Compressed Sensing MRI Reconstruction}.
\newblock IEEE Transactions on Medical Imaging (2018)

\bibitem{Ben-Cohen2017}
Ben-Cohen, A., Klang, E., Raskin, S.P., Amitai, M.M., Greenspan, H.:
\newblock {Virtual PET Images from CT Data Using Deep Convolutional Networks:
  Initial Results}.
\newblock  In: MICCAI Workshop on Simulation and Synthesis in Medical Imaging
  (2017)

\bibitem{Bayramolu2017}
Bayramolu, N., Kaakinen, M., Eklund, L.:
\newblock {Towards Virtual H{\&}E Staining of Hyperspectral Lung Histology
  Images Using Conditional Generative Adversarial Networks}.
\newblock  In: International Conference on Computer Vision (2017)

\bibitem{Chu2017}
Chu, C., Zhmoginov, A., Sandler, M.:
\newblock {CycleGAN, a Master of Steganography}.
\newblock  In: Neural Information Processing Systems Workshop on Machine
  Deception (2017)

\bibitem{mirza2014conditional}
Mirza, M., Osindero, S.:
\newblock {Conditional Generative Adversarial Nets}.
\newblock arXiv: 1411.1784 (2014)

\bibitem{Menze_et_al_2015}
Menze, B.H., Jakab, A., Bauer, S.,  et~al.:
\newblock {The Multimodal Brain Tumor Image Segmentation Benchmark (BRATS)}.
\newblock IEEE Transactions on Medical Imaging (2015)

\end{thebibliography}
}

\setcounter{figure}{0}
\renewcommand{\thefigure}{S\arabic{figure}}

\begin{figure}
    \scalebox{1}{{\centering
\textbf{\LARGE Supplementary Information}}} \\
\vspace{25pt}
    \captionsetup[subfigure]{labelformat=empty}
    \centering
    
    \begin{subfigure}[b]{1.0\textwidth}
    \caption{(a) Healthy examples  from the holdout test set}
    
    \begin{subfigure}[b]{0.07\textwidth}\includegraphics[width=\linewidth, trim={0 50 90 50}, clip]{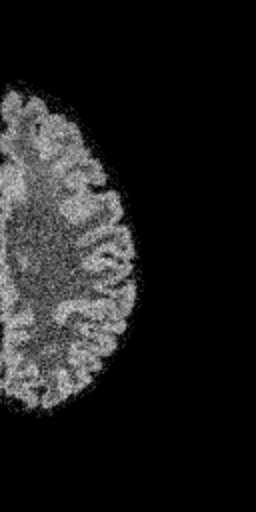}\end{subfigure}\hspace{10pt}%
    \begin{subfigure}[b]{0.07\textwidth}\includegraphics[width=\linewidth, trim={0 50 90 50}, clip]{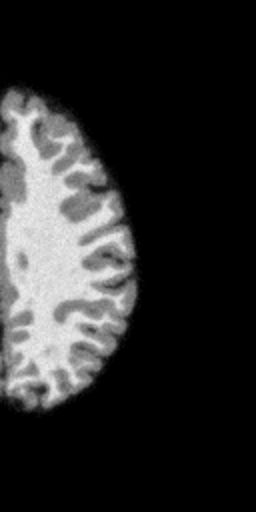}\end{subfigure}%
    \begin{subfigure}[b]{0.07\textwidth}\includegraphics[width=\linewidth, trim={0 50 90 50}, clip]{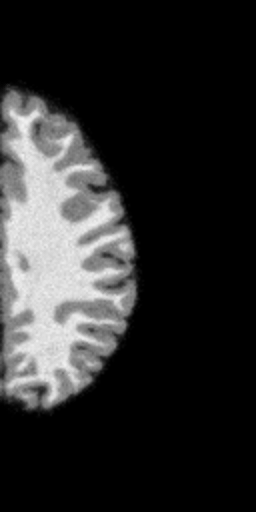}\end{subfigure}%
    \begin{subfigure}[b]{0.07\textwidth}\includegraphics[width=\linewidth, trim={0 50 90 50}, clip]{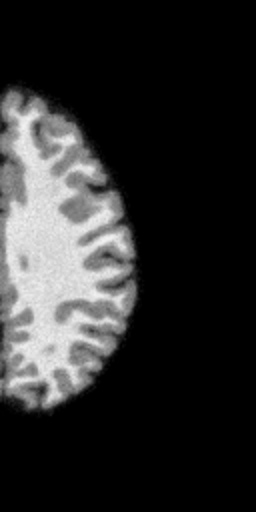}\end{subfigure}%
    \begin{subfigure}[b]{0.07\textwidth}\includegraphics[width=\linewidth, trim={0 50 90 50}, clip]{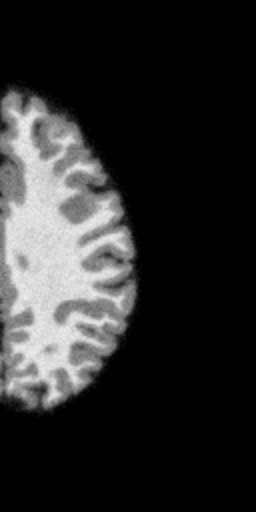}\end{subfigure}%
    \begin{subfigure}[b]{0.07\textwidth}\includegraphics[width=\linewidth, trim={0 50 90 50}, clip]{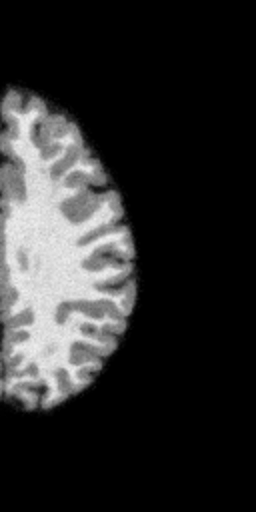}\end{subfigure}%
    \begin{subfigure}[b]{0.07\textwidth}\includegraphics[width=\linewidth, trim={0 50 90 50}, clip]{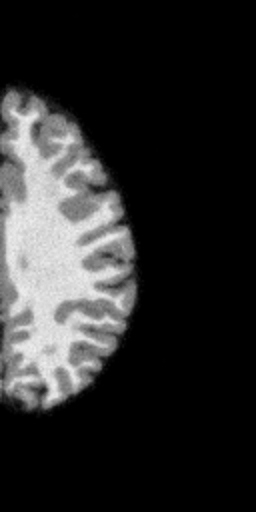}\end{subfigure}%
    \begin{subfigure}[b]{0.07\textwidth}\includegraphics[width=\linewidth, trim={0 50 90 50}, clip]{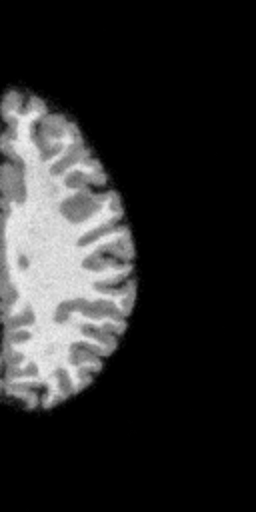}\end{subfigure}%
    \begin{subfigure}[b]{0.07\textwidth}\includegraphics[width=\linewidth, trim={0 50 90 50}, clip]{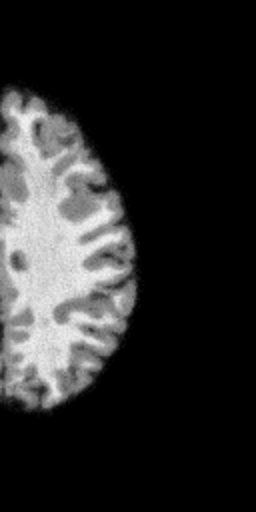}\end{subfigure}%
    \begin{subfigure}[b]{0.07\textwidth}\includegraphics[width=\linewidth, trim={0 50 90 50}, clip]{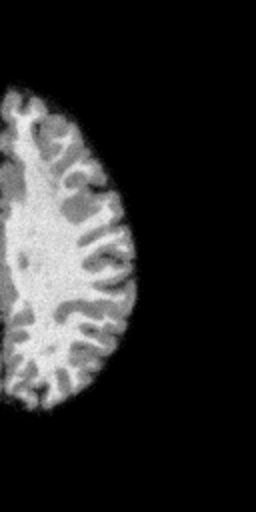}\end{subfigure}%
    \begin{subfigure}[b]{0.07\textwidth}\includegraphics[width=\linewidth, trim={0 50 90 50}, clip]{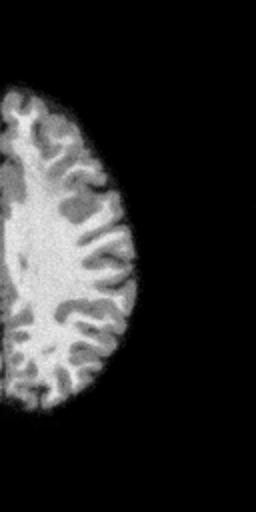}\end{subfigure}%
    \begin{subfigure}[b]{0.07\textwidth}\includegraphics[width=\linewidth, trim={0 50 90 50}, clip]{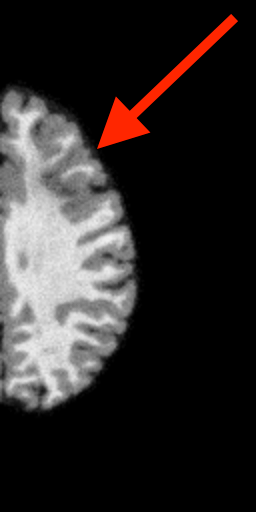}\end{subfigure}%
    \hspace{10pt}\begin{subfigure}[b]{0.07\textwidth}\includegraphics[width=\linewidth, trim={0 50 90 50}, clip]{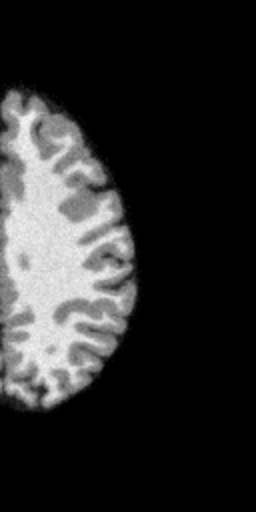}\end{subfigure}%
    
    \begin{subfigure}[b]{0.07\textwidth}\includegraphics[width=\linewidth, trim={0 50 90 50}, clip]{Figs/LG0012-114-False_real_A.png}\end{subfigure}\hspace{10pt}%
    \begin{subfigure}[b]{0.07\textwidth}\includegraphics[width=\linewidth, trim={0 50 90 50}, clip]{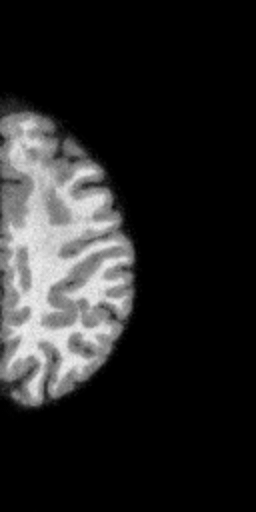}\end{subfigure}%
    \begin{subfigure}[b]{0.07\textwidth}\includegraphics[width=\linewidth, trim={0 50 90 50}, clip]{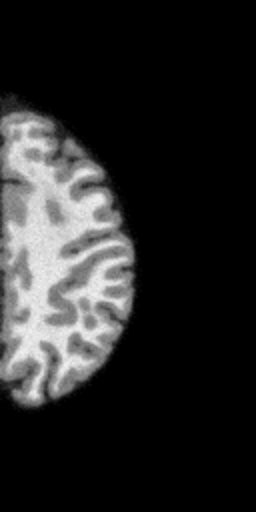}\end{subfigure}%
    \begin{subfigure}[b]{0.07\textwidth}\includegraphics[width=\linewidth, trim={0 50 90 50}, clip]{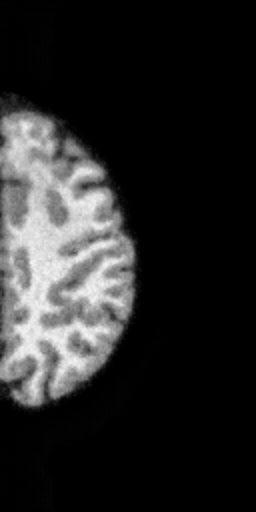}\end{subfigure}%
    \begin{subfigure}[b]{0.07\textwidth}\includegraphics[width=\linewidth, trim={0 50 90 50}, clip]{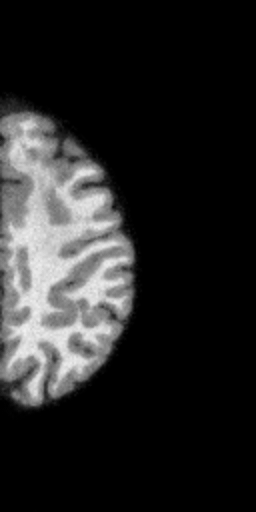}\end{subfigure}%
    \begin{subfigure}[b]{0.07\textwidth}\includegraphics[width=\linewidth, trim={0 50 90 50}, clip]{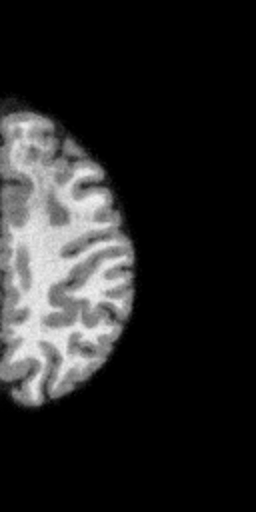}\end{subfigure}%
    \begin{subfigure}[b]{0.07\textwidth}\includegraphics[width=\linewidth, trim={0 50 90 50}, clip]{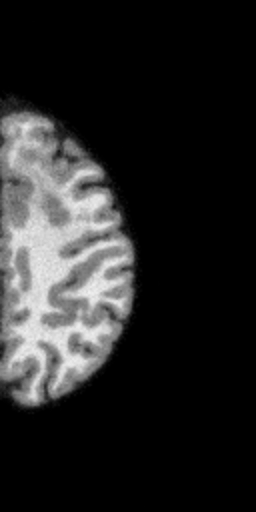}\end{subfigure}%
    \begin{subfigure}[b]{0.07\textwidth}\includegraphics[width=\linewidth, trim={0 50 90 50}, clip]{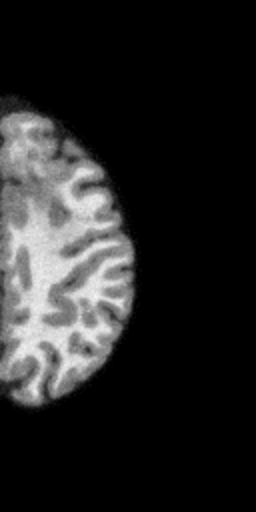}\end{subfigure}%
    \begin{subfigure}[b]{0.07\textwidth}\includegraphics[width=\linewidth, trim={0 50 90 50}, clip]{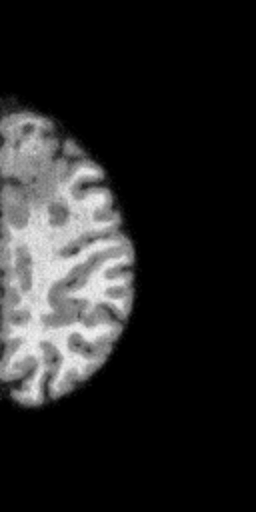}\end{subfigure}%
    \begin{subfigure}[b]{0.07\textwidth}\includegraphics[width=\linewidth, trim={0 50 90 50}, clip]{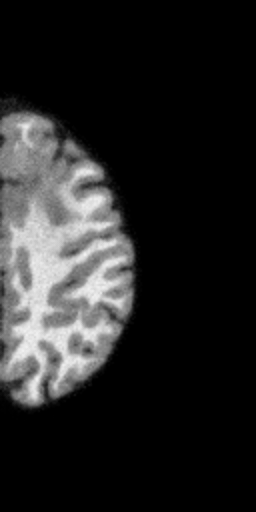}\end{subfigure}%
    \begin{subfigure}[b]{0.07\textwidth}\includegraphics[width=\linewidth, trim={0 50 90 50}, clip]{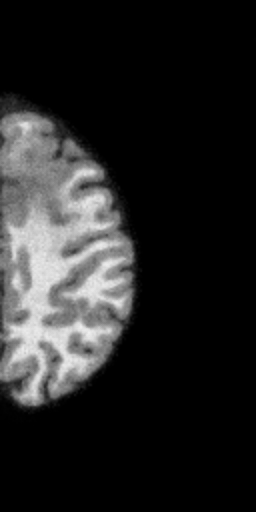}\end{subfigure}%
    \begin{subfigure}[b]{0.07\textwidth}\includegraphics[width=\linewidth, trim={0 50 90 50}, clip]{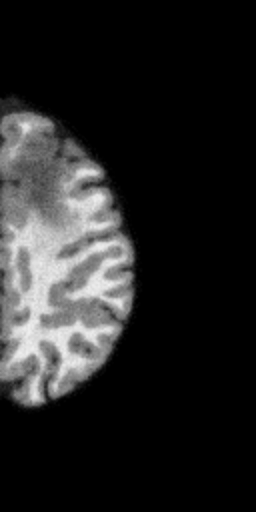}\end{subfigure}%
    \hspace{10pt}\begin{subfigure}[b]{0.07\textwidth}\includegraphics[width=\linewidth, trim={0 50 90 50}, clip]{Figs/LG0012-114-False_real_B.png}\end{subfigure}%
    
    \begin{subfigure}[b]{0.07\textwidth}\includegraphics[width=\linewidth, trim={0 50 90 50}, clip]{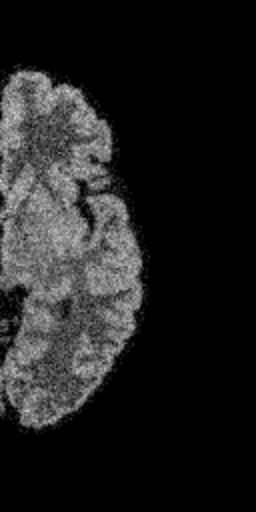}\end{subfigure}\hspace{10pt}%
    \begin{subfigure}[b]{0.07\textwidth}\includegraphics[width=\linewidth, trim={0 50 90 50}, clip]{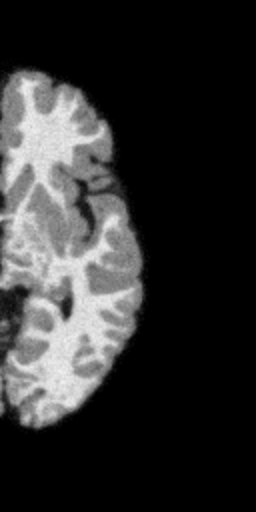}\end{subfigure}%
    \begin{subfigure}[b]{0.07\textwidth}\includegraphics[width=\linewidth, trim={0 50 90 50}, clip]{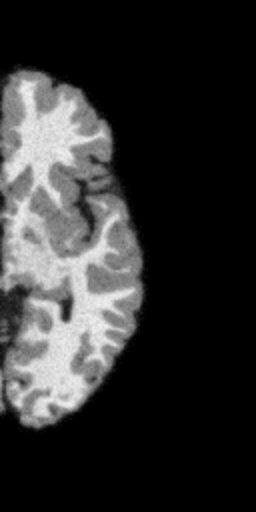}\end{subfigure}%
    \begin{subfigure}[b]{0.07\textwidth}\includegraphics[width=\linewidth, trim={0 50 90 50}, clip]{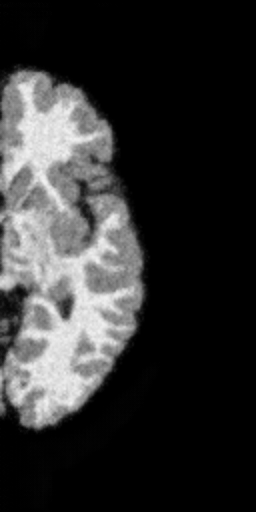}\end{subfigure}%
    \begin{subfigure}[b]{0.07\textwidth}\includegraphics[width=\linewidth, trim={0 50 90 50}, clip]{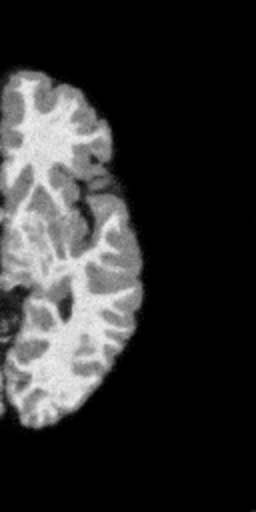}\end{subfigure}%
    \begin{subfigure}[b]{0.07\textwidth}\includegraphics[width=\linewidth, trim={0 50 90 50}, clip]{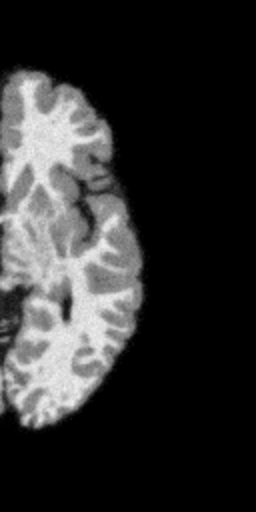}\end{subfigure}%
    \begin{subfigure}[b]{0.07\textwidth}\includegraphics[width=\linewidth, trim={0 50 90 50}, clip]{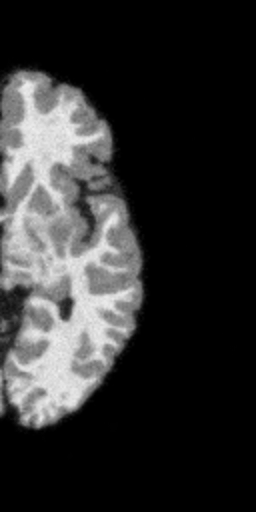}\end{subfigure}%
    \begin{subfigure}[b]{0.07\textwidth}\includegraphics[width=\linewidth, trim={0 50 90 50}, clip]{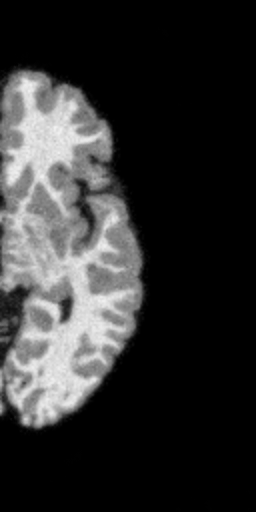}\end{subfigure}%
    \begin{subfigure}[b]{0.07\textwidth}\includegraphics[width=\linewidth, trim={0 50 90 50}, clip]{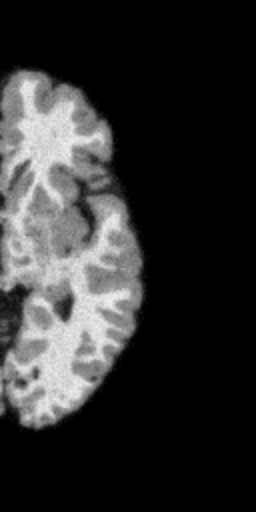}\end{subfigure}%
    \begin{subfigure}[b]{0.07\textwidth}\includegraphics[width=\linewidth, trim={0 50 90 50}, clip]{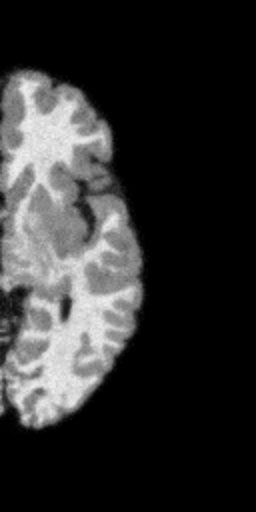}\end{subfigure}%
    \begin{subfigure}[b]{0.07\textwidth}\includegraphics[width=\linewidth, trim={0 50 90 50}, clip]{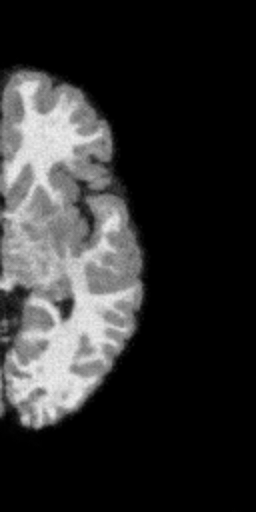}\end{subfigure}%
    \begin{subfigure}[b]{0.07\textwidth}\includegraphics[width=\linewidth, trim={0 50 90 50}, clip]{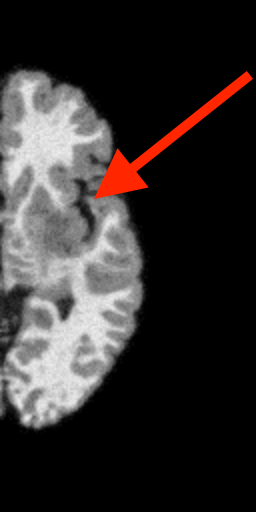}\end{subfigure}%
    \hspace{10pt}\begin{subfigure}[b]{0.07\textwidth}\includegraphics[width=\linewidth, trim={0 50 90 50}, clip]{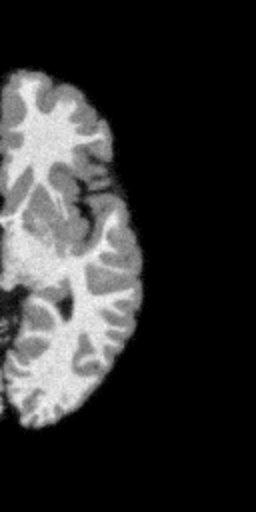}\end{subfigure}%
    \end{subfigure}%
 
    \begin{subfigure}[b]{1.0\textwidth}
    \caption{\\ (b) Examples with a tumor  from the holdout test set}

    \begin{subfigure}[b]{0.07\textwidth}\includegraphics[width=\linewidth, trim={0 50 90 50}, clip]{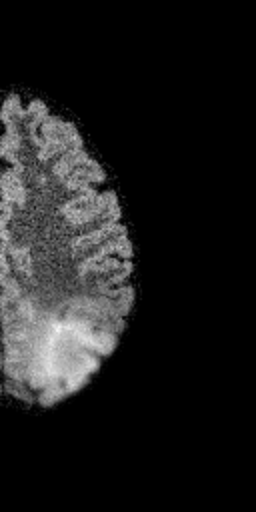}\end{subfigure}\hspace{10pt}%
    \begin{subfigure}[b]{0.07\textwidth}\includegraphics[width=\linewidth, trim={0 50 90 50}, clip]{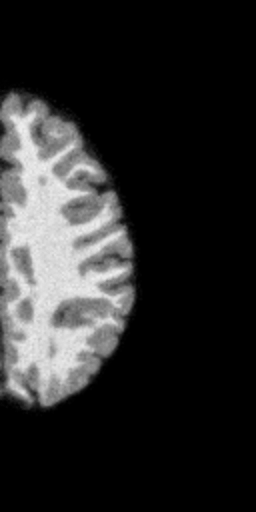}\end{subfigure}%
    \begin{subfigure}[b]{0.07\textwidth}\includegraphics[width=\linewidth, trim={0 50 90 50}, clip]{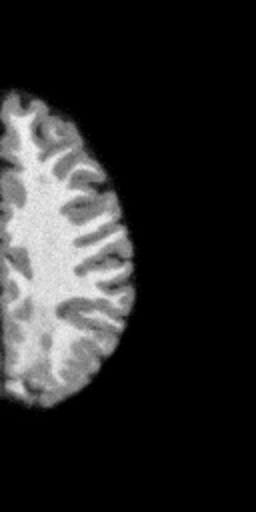}\end{subfigure}%
    \begin{subfigure}[b]{0.07\textwidth}\includegraphics[width=\linewidth, trim={0 50 90 50}, clip]{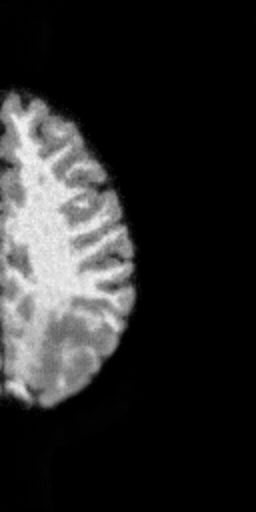}\end{subfigure}%
    \begin{subfigure}[b]{0.07\textwidth}\includegraphics[width=\linewidth, trim={0 50 90 50}, clip]{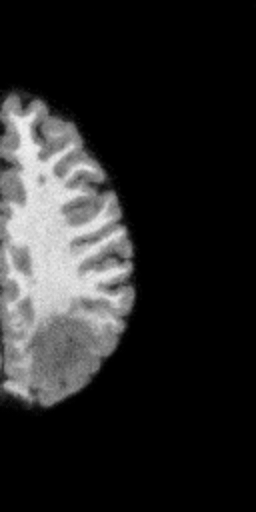}\end{subfigure}%
    \begin{subfigure}[b]{0.07\textwidth}\includegraphics[width=\linewidth, trim={0 50 90 50}, clip]{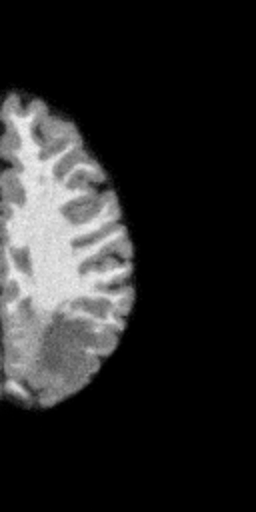}\end{subfigure}%
    \begin{subfigure}[b]{0.07\textwidth}\includegraphics[width=\linewidth, trim={0 50 90 50}, clip]{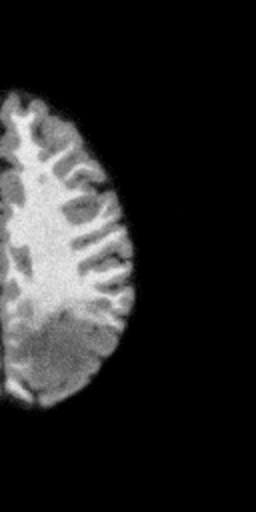}\end{subfigure}%
    \begin{subfigure}[b]{0.07\textwidth}\includegraphics[width=\linewidth, trim={0 50 90 50}, clip]{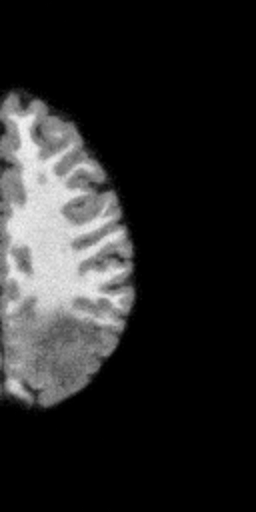}\end{subfigure}%
    \begin{subfigure}[b]{0.07\textwidth}\includegraphics[width=\linewidth, trim={0 50 90 50}, clip]{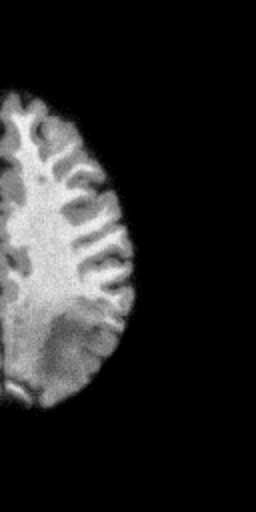}\end{subfigure}%
    \begin{subfigure}[b]{0.07\textwidth}\includegraphics[width=\linewidth, trim={0 50 90 50}, clip]{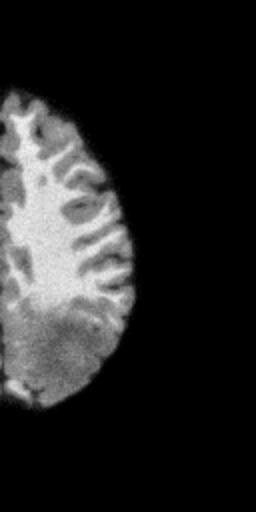}\end{subfigure}%
    \begin{subfigure}[b]{0.07\textwidth}\includegraphics[width=\linewidth, trim={0 50 90 50}, clip]{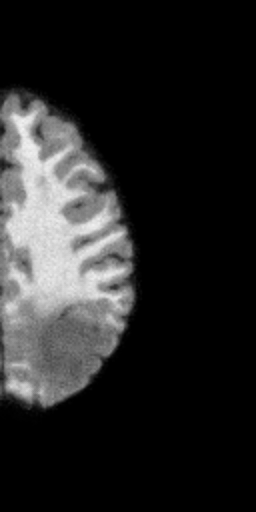}\end{subfigure}%
    \begin{subfigure}[b]{0.07\textwidth}\includegraphics[width=\linewidth, trim={0 50 90 50}, clip]{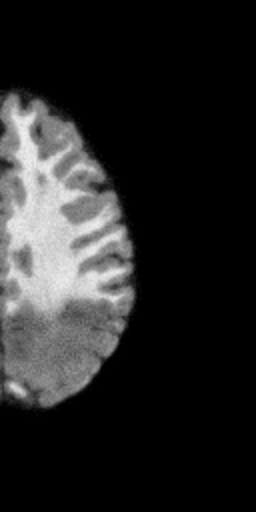}\end{subfigure}%
    \hspace{10pt}\begin{subfigure}[b]{0.07\textwidth}\includegraphics[width=\linewidth, trim={0 50 90 50}, clip]{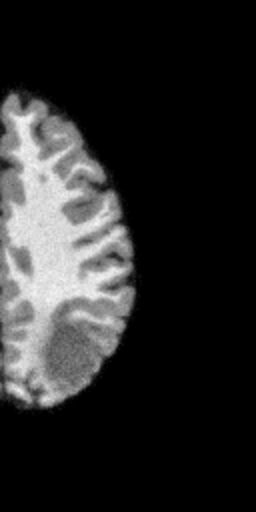}\end{subfigure}%

    \begin{subfigure}[b]{0.07\textwidth}\includegraphics[width=\linewidth, trim={0 50 90 50}, clip]{Figs/HG0001-103-True_real_A.png}\end{subfigure}\hspace{10pt}%
    \begin{subfigure}[b]{0.07\textwidth}\includegraphics[width=\linewidth, trim={0 50 90 50}, clip]{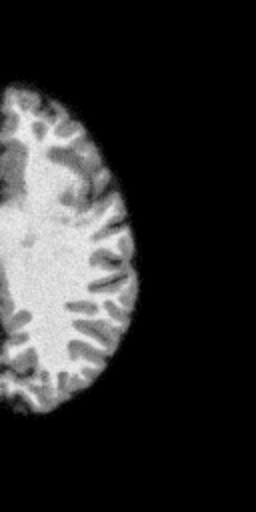}\end{subfigure}%
    \begin{subfigure}[b]{0.07\textwidth}\includegraphics[width=\linewidth, trim={0 50 90 50}, clip]{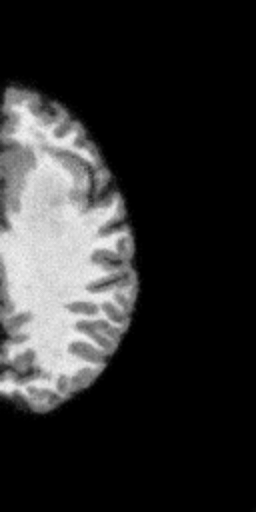}\end{subfigure}%
    \begin{subfigure}[b]{0.07\textwidth}\includegraphics[width=\linewidth, trim={0 50 90 50}, clip]{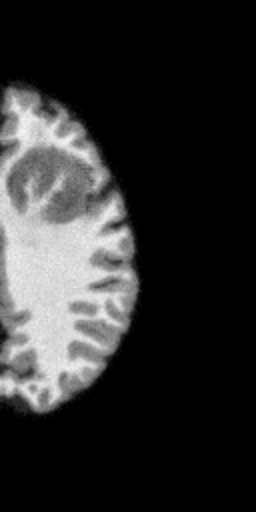}\end{subfigure}%
    \begin{subfigure}[b]{0.07\textwidth}\includegraphics[width=\linewidth, trim={0 50 90 50}, clip]{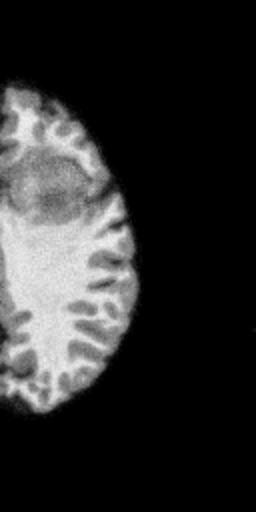}\end{subfigure}%
    \begin{subfigure}[b]{0.07\textwidth}\includegraphics[width=\linewidth, trim={0 50 90 50}, clip]{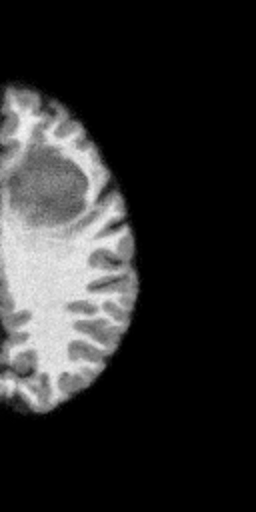}\end{subfigure}%
    \begin{subfigure}[b]{0.07\textwidth}\includegraphics[width=\linewidth, trim={0 50 90 50}, clip]{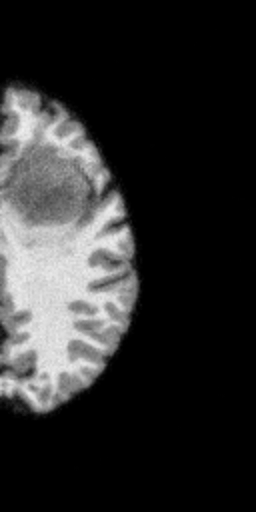}\end{subfigure}%
    \begin{subfigure}[b]{0.07\textwidth}\includegraphics[width=\linewidth, trim={0 50 90 50}, clip]{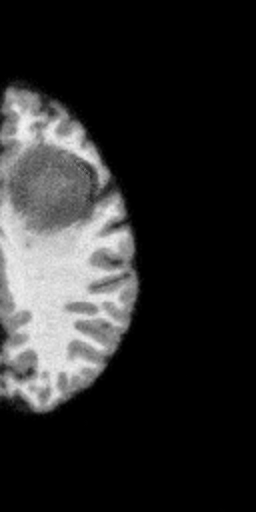}\end{subfigure}%
    \begin{subfigure}[b]{0.07\textwidth}\includegraphics[width=\linewidth, trim={0 50 90 50}, clip]{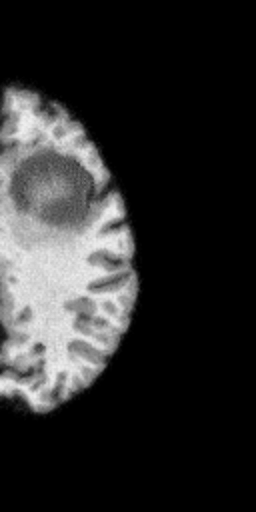}\end{subfigure}%
    \begin{subfigure}[b]{0.07\textwidth}\includegraphics[width=\linewidth, trim={0 50 90 50}, clip]{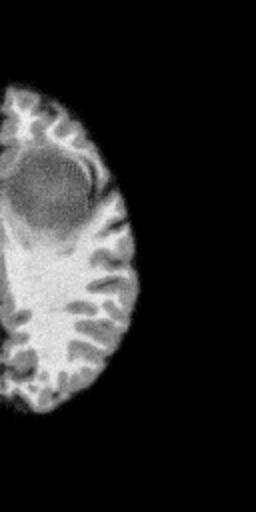}\end{subfigure}%
    \begin{subfigure}[b]{0.07\textwidth}\includegraphics[width=\linewidth, trim={0 50 90 50}, clip]{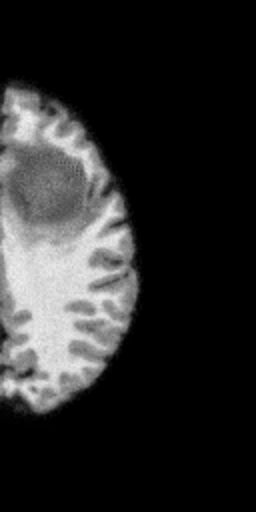}\end{subfigure}%
    \begin{subfigure}[b]{0.07\textwidth}\includegraphics[width=\linewidth, trim={0 50 90 50}, clip]{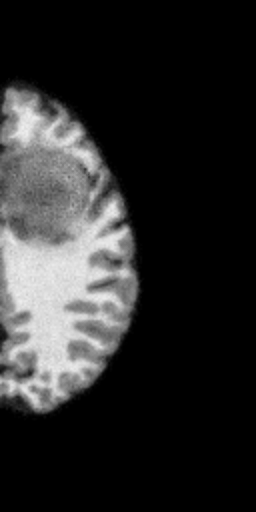}\end{subfigure}%
    \hspace{10pt}\begin{subfigure}[b]{0.07\textwidth}\includegraphics[width=\linewidth, trim={0 50 90 50}, clip]{Figs/HG0001-103-True_real_B.png}\end{subfigure}%

    \begin{subfigure}[b]{0.07\textwidth}\includegraphics[width=\linewidth, trim={0 50 90 50}, clip]{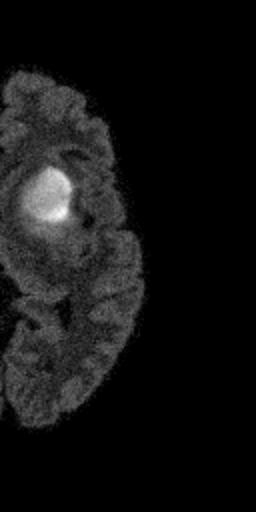}\caption{\centering \begin{minipage}{1\textwidth}\centering \vspace{2 px} Flair \\ (source) \end{minipage}}\end{subfigure}\hspace{10pt}%
    \begin{subfigure}[b]{0.07\textwidth}\includegraphics[width=\linewidth, trim={0 50 90 50}, clip]{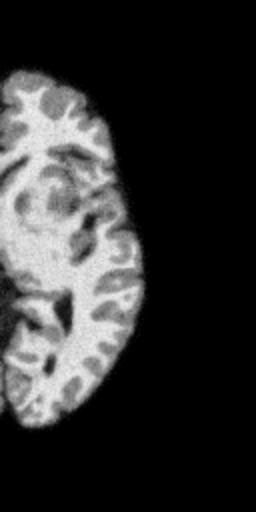}\caption{0\%\\\hspace{1pt}}\end{subfigure}%
    \begin{subfigure}[b]{0.07\textwidth}\includegraphics[width=\linewidth, trim={0 50 90 50}, clip]{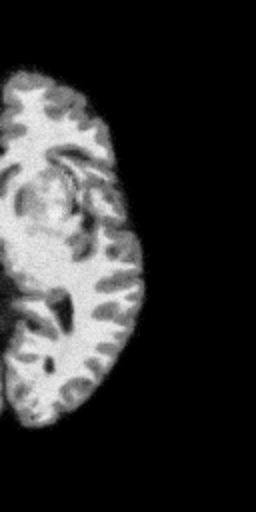}\caption{10\%\\\hspace{1pt}}\end{subfigure}%
    \begin{subfigure}[b]{0.07\textwidth}\includegraphics[width=\linewidth, trim={0 50 90 50}, clip]{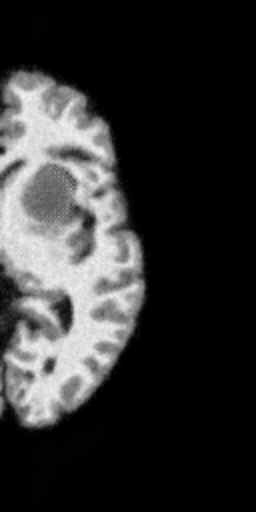}\caption{20\%\\\hspace{1pt}}\end{subfigure}%
    \begin{subfigure}[b]{0.07\textwidth}\includegraphics[width=\linewidth, trim={0 50 90 50}, clip]{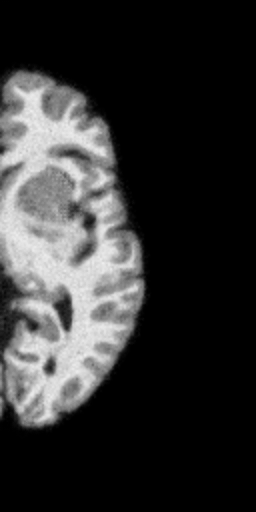}\caption{30\%\\\hspace{1pt}}\end{subfigure}%
    \begin{subfigure}[b]{0.07\textwidth}\includegraphics[width=\linewidth, trim={0 50 90 50}, clip]{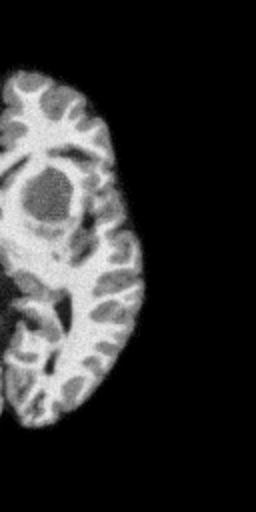}\caption{40\%\\\hspace{1pt}}\end{subfigure}%
    \begin{subfigure}[b]{0.07\textwidth}\includegraphics[width=\linewidth, trim={0 50 90 50}, clip]{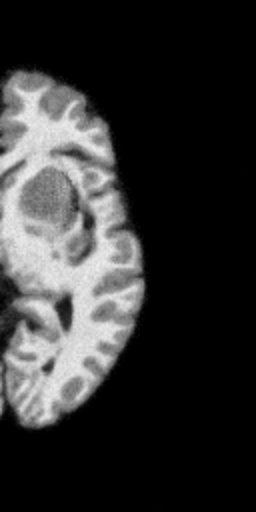}\caption{50\%\\\hspace{1pt}}\end{subfigure}%
    \begin{subfigure}[b]{0.07\textwidth}\includegraphics[width=\linewidth, trim={0 50 90 50}, clip]{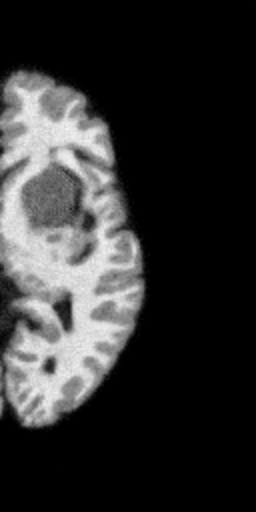}\caption{60\%\\\hspace{1pt}}\end{subfigure}%
    \begin{subfigure}[b]{0.07\textwidth}\includegraphics[width=\linewidth, trim={0 50 90 50}, clip]{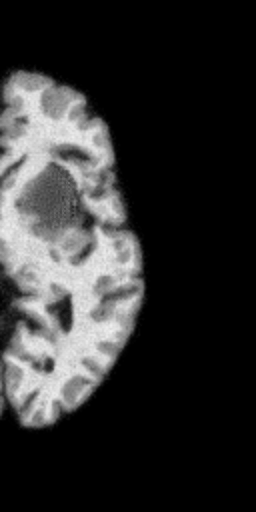}\caption{70\%\\\hspace{1pt}}\end{subfigure}%
    \begin{subfigure}[b]{0.07\textwidth}\includegraphics[width=\linewidth, trim={0 50 90 50}, clip]{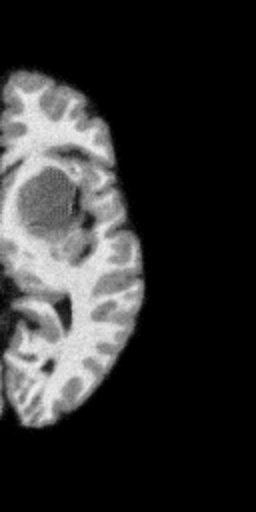}\caption{80\%\\\hspace{1pt}}\end{subfigure}%
    \begin{subfigure}[b]{0.07\textwidth}\includegraphics[width=\linewidth, trim={0 50 90 50}, clip]{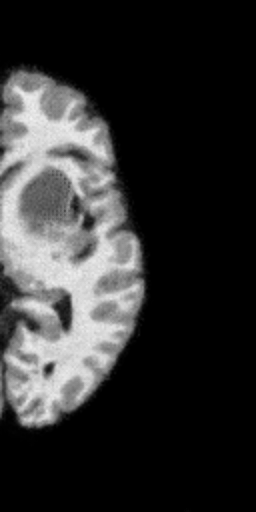}\caption{90\%\\\hspace{1pt}}\end{subfigure}%
    \begin{subfigure}[b]{0.07\textwidth}\includegraphics[width=\linewidth, trim={0 50 90 50}, clip]{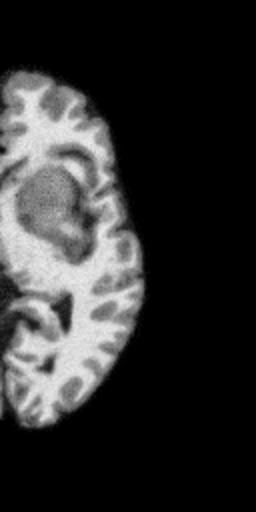}\caption{100\%\\\hspace{1pt}}\end{subfigure}%
    \hspace{10pt}\begin{subfigure}[b]{0.07\textwidth}\includegraphics[width=\linewidth, trim={0 50 90 50}, clip]{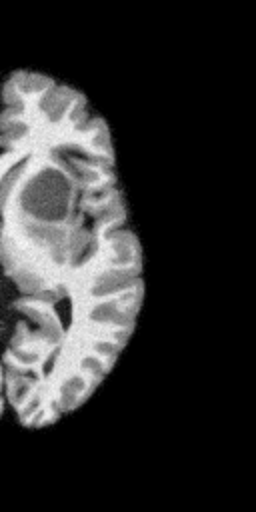}\caption{\centering \begin{minipage}{1\textwidth}\vspace{2 px}\centering{T1\\(target)} \end{minipage}}\end{subfigure}%
    \end{subfigure}%

    \caption{Illustration of healthy (a) and tumor (b) class change through domain translation while changing the ratio of the healthy to tumor samples in the target domain $D_b$ for CycleGAN. We vary the distribution of $D_b$ from 0\% tumor examples to 100\% tumor examples to train 11 different CycleGAN models. On the left you see the image in the source domain (Flair) and on the right you see the corresponding ground truth image in the target domain (T1). The {\color{red}red} arrows point out hard to spot tumors that were introduced.}
    \label{fig:varyb}
\end{figure}

\begin{figure}
\begin{center}
\centering
\begin{subfigure}[b]{1.0\textwidth}
\caption{\small When source image is healthy and classifier predicts healthy}
\scalebox{.4}{{\hspace{17pt} (Source) \hspace{344pt} (target) \hspace{2pt} (Source) \hspace{344pt} (target)}} \\
\scalebox{.6}{\rotatebox{90}{\hspace{21pt} L1 \hspace{21pt} CondGAN \hspace{6pt} CycleGAN}}
{\includegraphics[width=0.48\textwidth]{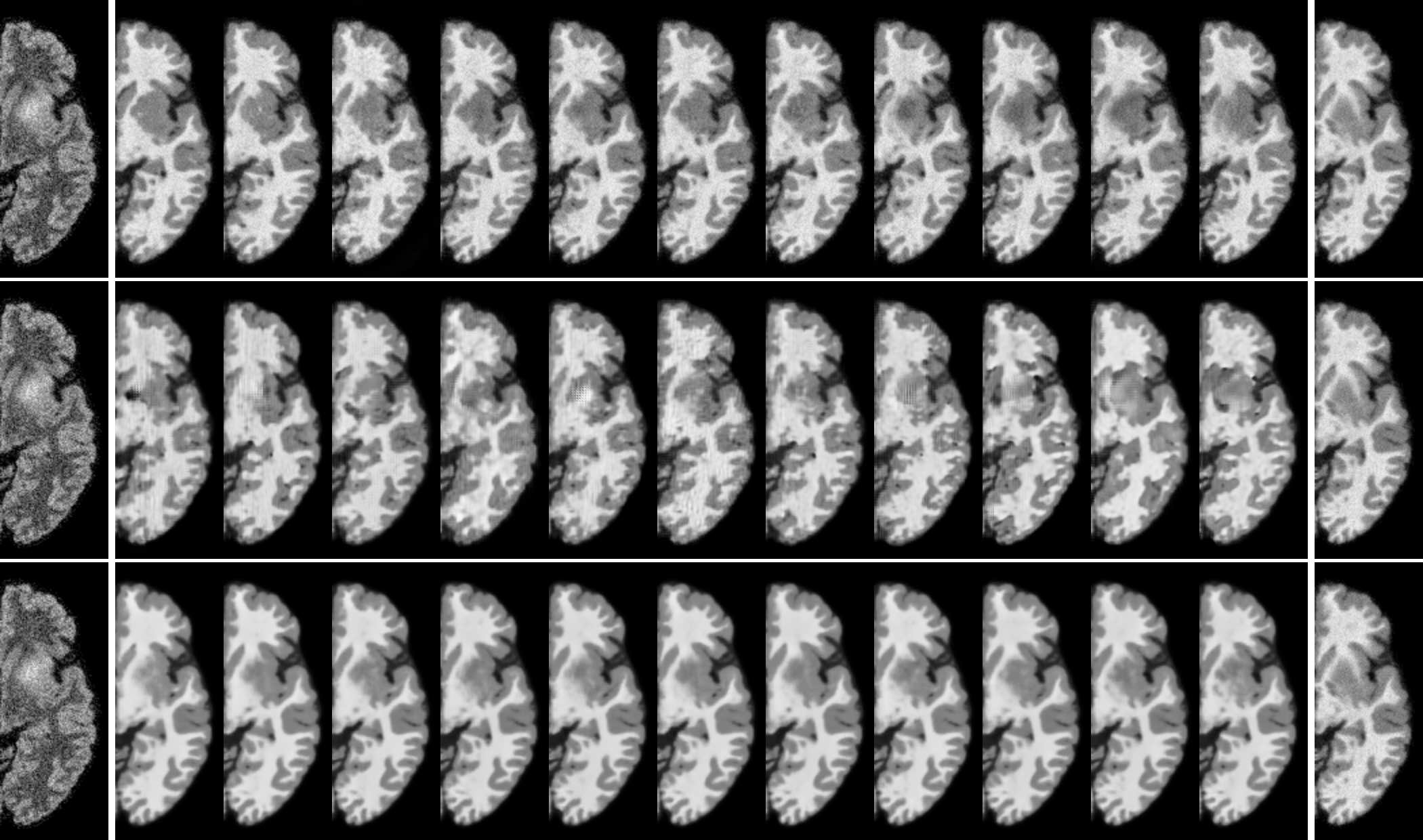}}
\fboxsep=0mm
\fboxrule=1pt
{\includegraphics[width=0.48\textwidth]{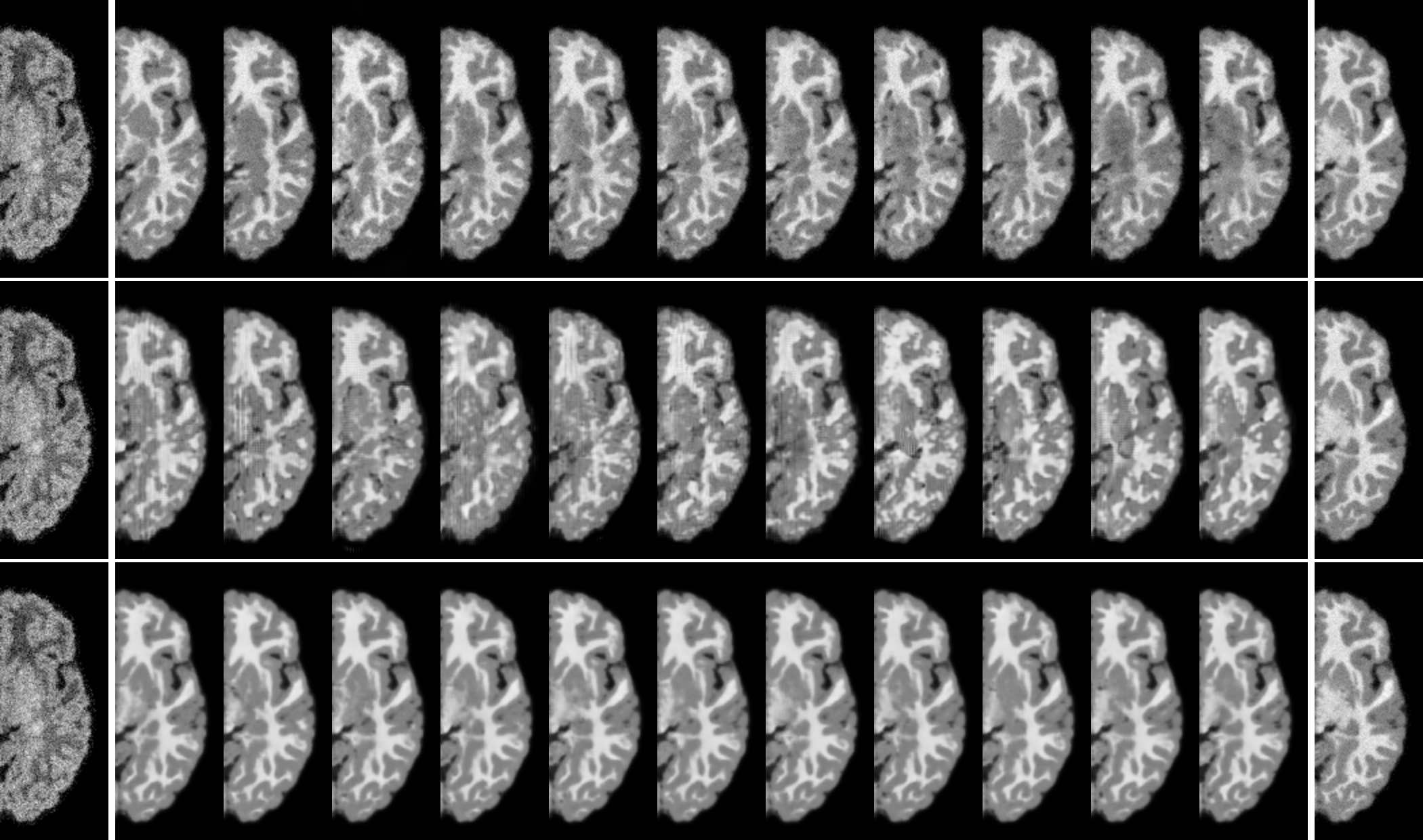}}
\end{subfigure} \\
\vskip6pt
\begin{subfigure}[b]{1.0\textwidth}
\caption{{\small When source image has tumor and classifier predicts tumor}}
\scalebox{.6}{\rotatebox{90}{\hspace{21pt} L1 \hspace{21pt} CondGAN \hspace{6pt} CycleGAN}}
{\includegraphics[width=0.48\textwidth]{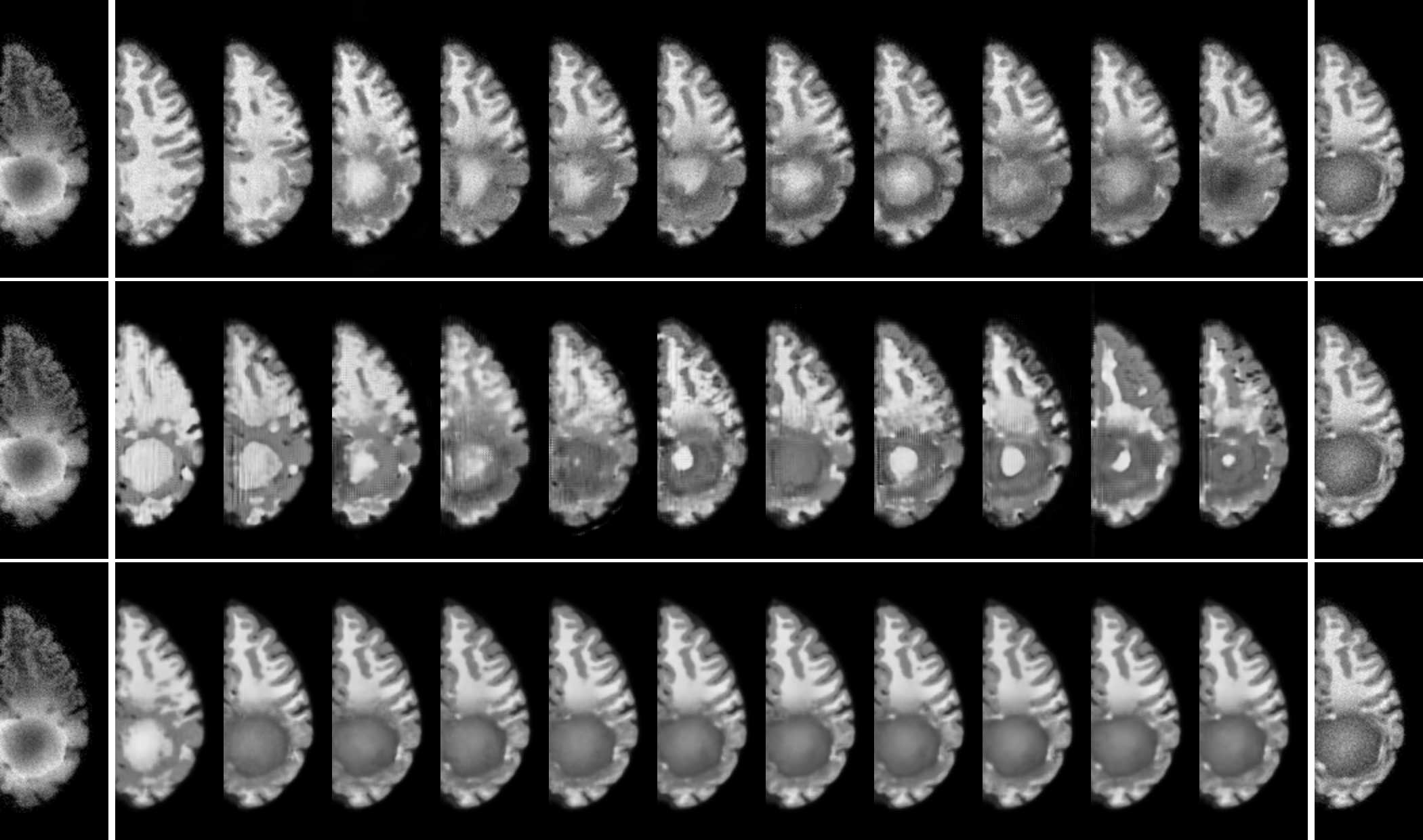}}
\fboxsep=0mm
\fboxrule=1pt
{\includegraphics[width=0.48\textwidth]{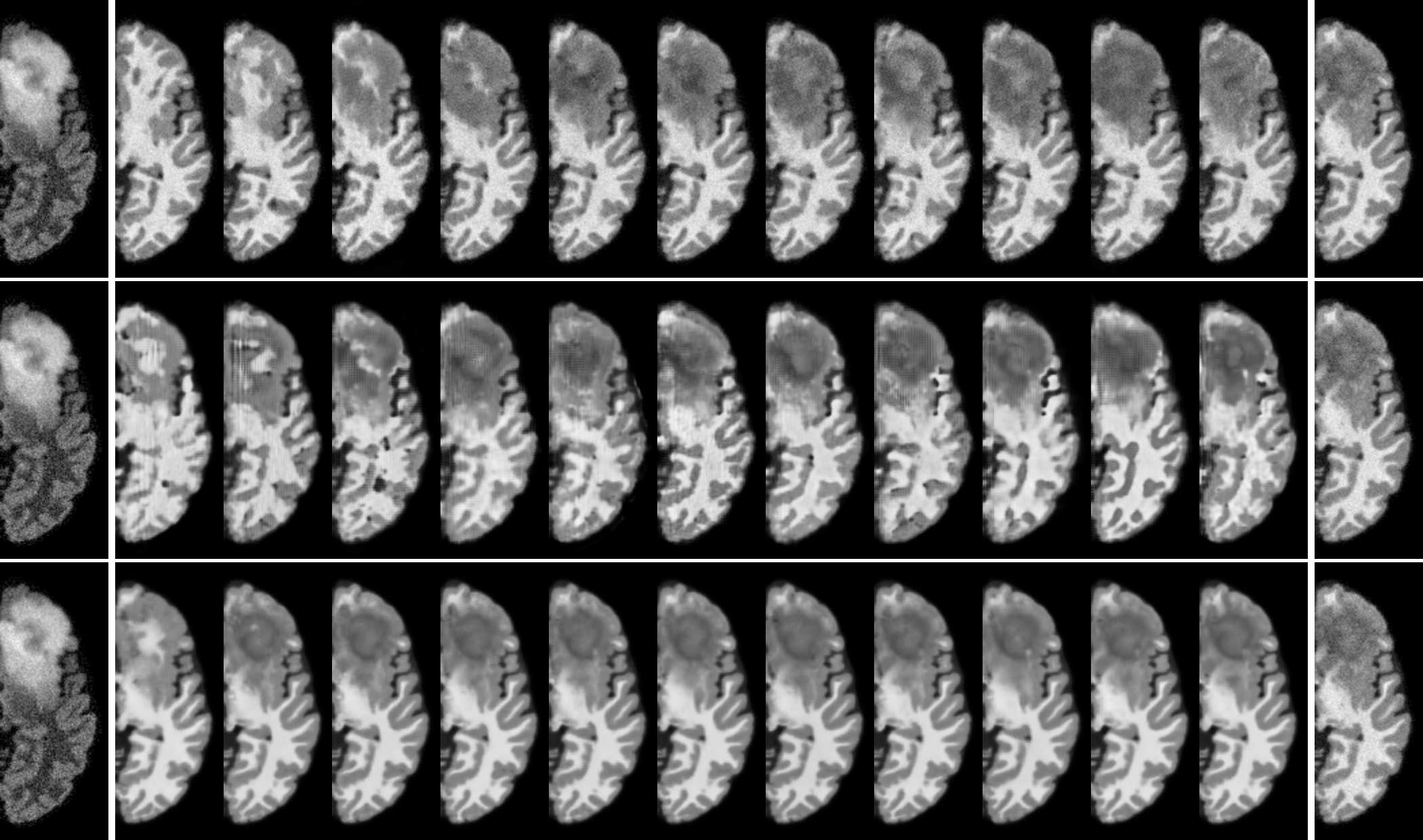}}
\vskip2pt
\scalebox{.6}{\rotatebox{90}{\hspace{21pt} L1 \hspace{21pt} CondGAN \hspace{6pt} CycleGAN}}
{\includegraphics[width=0.48\textwidth]{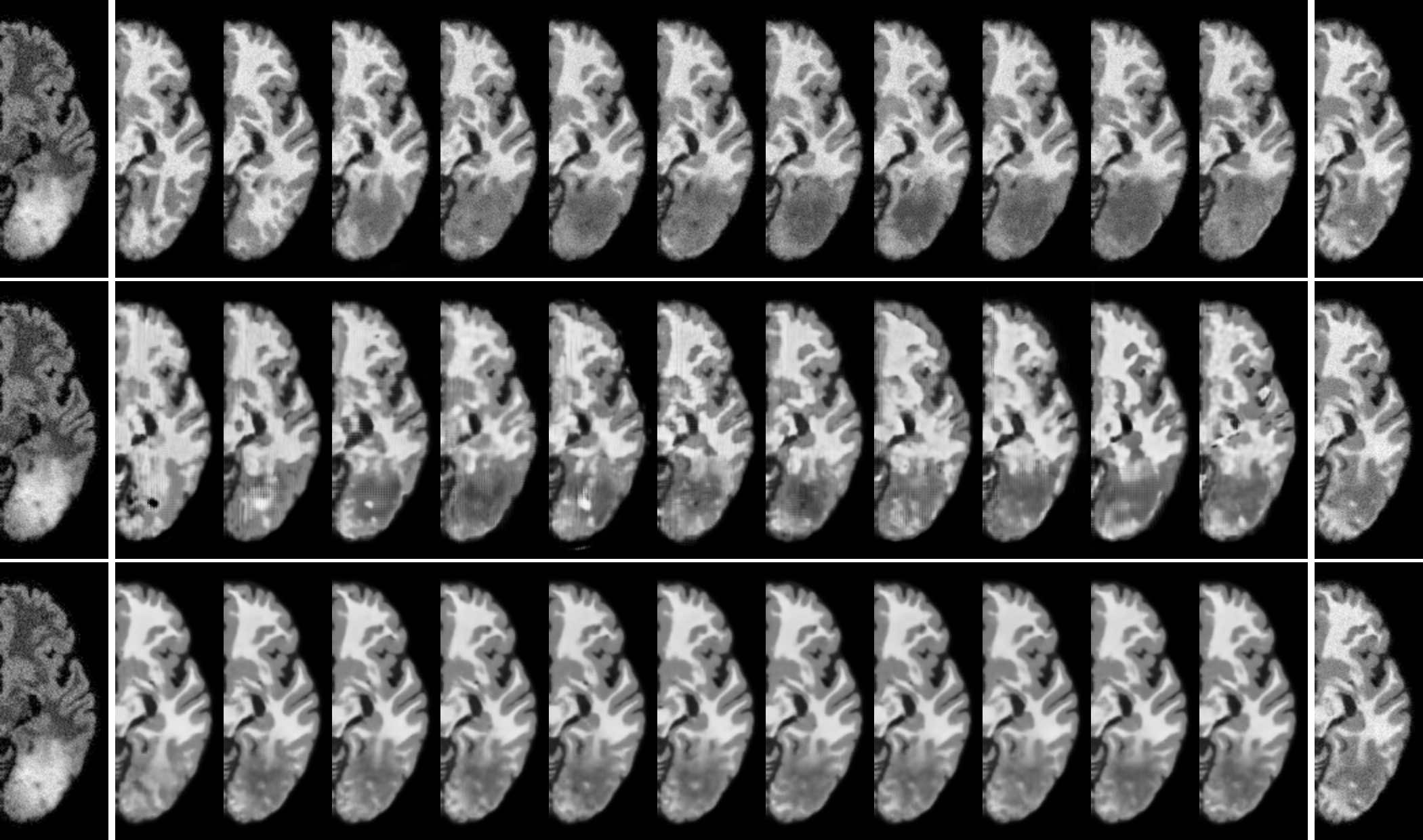}}
\fboxsep=0mm
\fboxrule=1pt
{\includegraphics[width=0.48\textwidth]{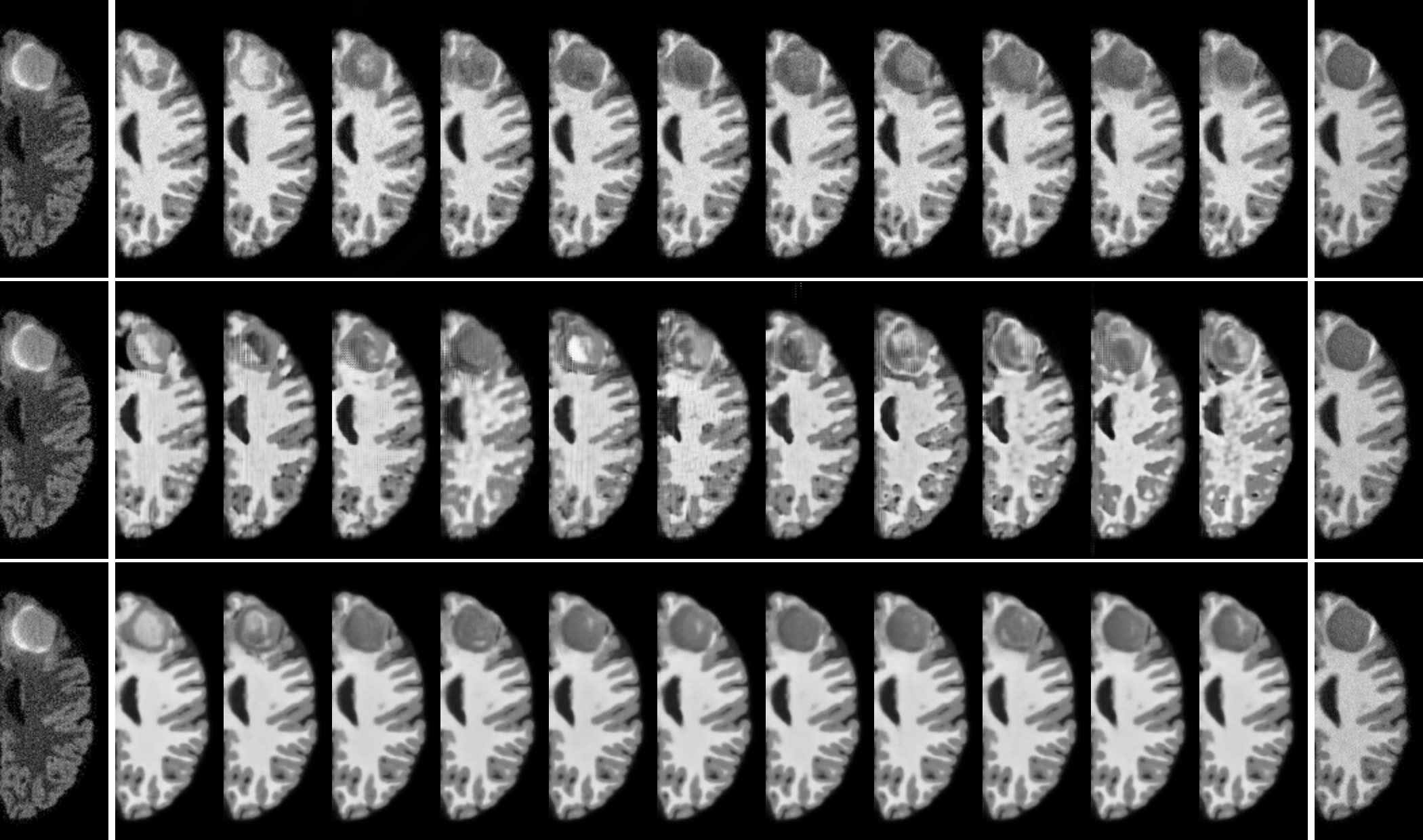}}\\
\fboxsep=0mm
\fboxrule=1pt
{\scalebox{.5}{{\hspace{15pt} Flair \hspace{4pt} 0\% \hspace{4pt} 10\% \hspace{4pt} 20\% \hspace{3.5pt} 30\% \hspace{3pt} 40\% \hspace{2.5pt} 50\% \hspace{2pt} 60\% \hspace{2pt} 70\% \hspace{2pt} 80\% \hspace{1pt} 90\% \hspace{1pt} 100\% \hspace{3pt} T1
\hspace{11pt} Flair \hspace{4pt} 0\% \hspace{4pt} 10\% \hspace{4pt} 20\% \hspace{3.5pt} 30\% \hspace{3pt} 40\% \hspace{2.5pt} 50\% \hspace{2pt} 60\% \hspace{2pt} 70\% \hspace{2pt} 80\% \hspace{1pt} 90\% \hspace{1pt} 100\% \hspace{3pt} T1}}}
\end{subfigure} 
\end{center}
\vspace{-5pt}
    \caption{\footnotesize 
    Showing translation bias when the classifier detects the class properly. On the top row, two examples are shown where the source image is healthy and the classifier predicts healthy for all 33 translated images across all 3 losses (CycleGAN, CondGAN, L1). Note how the translated images are still different from the ground truth T1 image in the target domain.
    On the bottom two rows, the source image has a tumor and the classifier predicts tumor for all 33 translated images. Note how the tumor seem to disappear in the images on the left and gets bigger in the images on the right. L1 loss suffers less from these artifacts, while adversarial losses produce more erroneous results. These samples show that even when the classifier is doing a perfect job, there is a bias in the translated images, and the bias is different for different losses.
    }
\label{fig:bias_class_same}
\end{figure}

\begin{figure}
\begin{center}
\centering
\begin{subfigure}[b]{1.0\textwidth}
\caption{\small When source image is healthy and classifier predicts tumor}
\scalebox{.4}{{\hspace{17pt} (Source) \hspace{344pt} (target) \hspace{2pt} (Source) \hspace{344pt} (target)}} \\
\scalebox{.6}{\rotatebox{90}{\hspace{21pt} L1 \hspace{21pt} CondGAN \hspace{6pt} CycleGAN}}
{\includegraphics[width=0.48\textwidth]{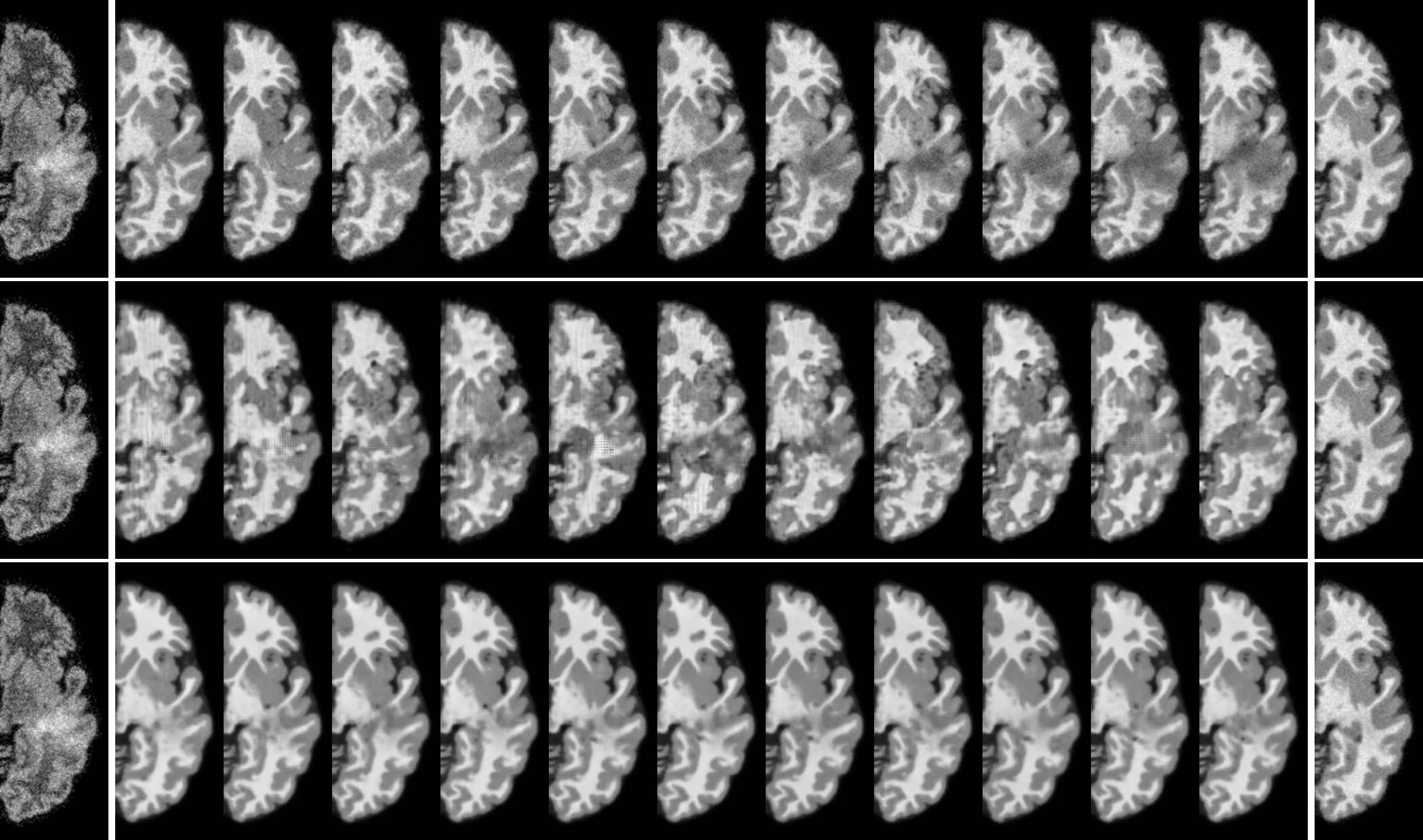}}
\fboxsep=0mm
\fboxrule=1pt
{\includegraphics[width=0.48\textwidth]{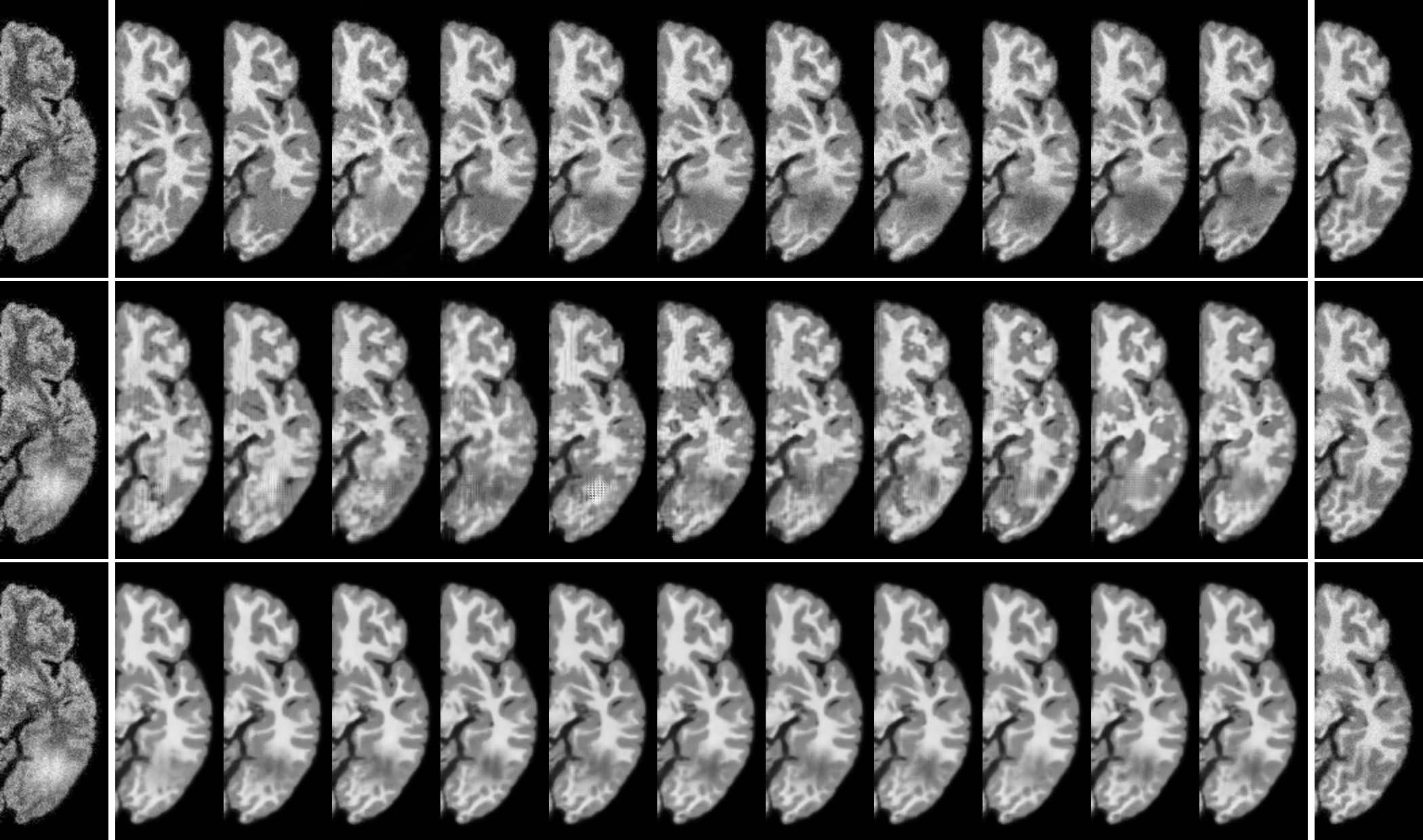}}
\end{subfigure}
\vskip6pt
\begin{subfigure}[b]{1.0\textwidth}
\caption{{\small When source image has tumor and classifier predicts healthy}}
\scalebox{.6}{\rotatebox{90}{\hspace{21pt} L1 \hspace{21pt} CondGAN \hspace{6pt} CycleGAN}}
{\includegraphics[width=0.48\textwidth]{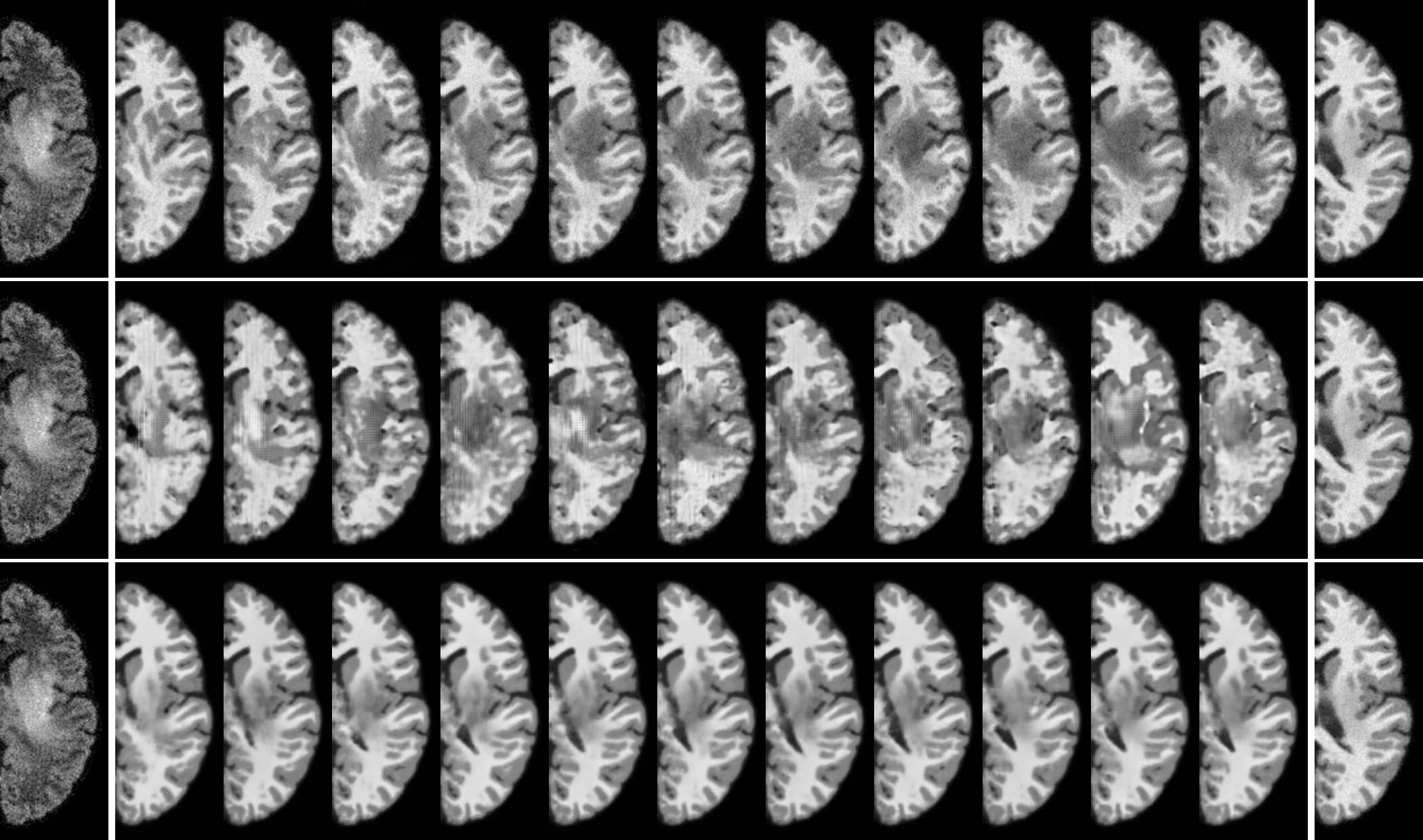}}
\fboxsep=0mm
\fboxrule=1pt
{\includegraphics[width=0.48\textwidth]{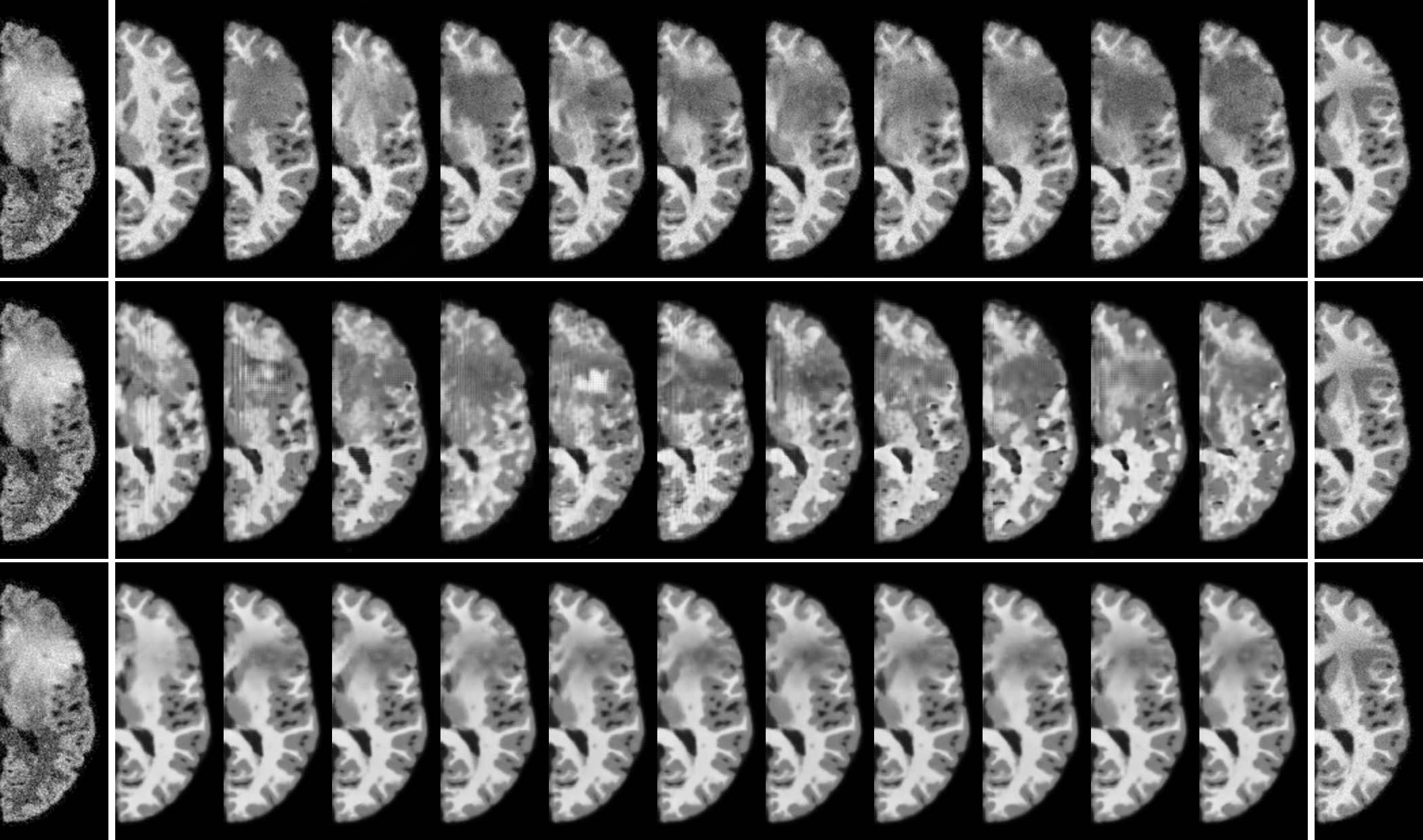}}
\vskip2pt
\scalebox{.6}{\rotatebox{90}{\hspace{21pt} L1 \hspace{21pt} CondGAN \hspace{6pt} CycleGAN}}
{\includegraphics[width=0.48\textwidth]{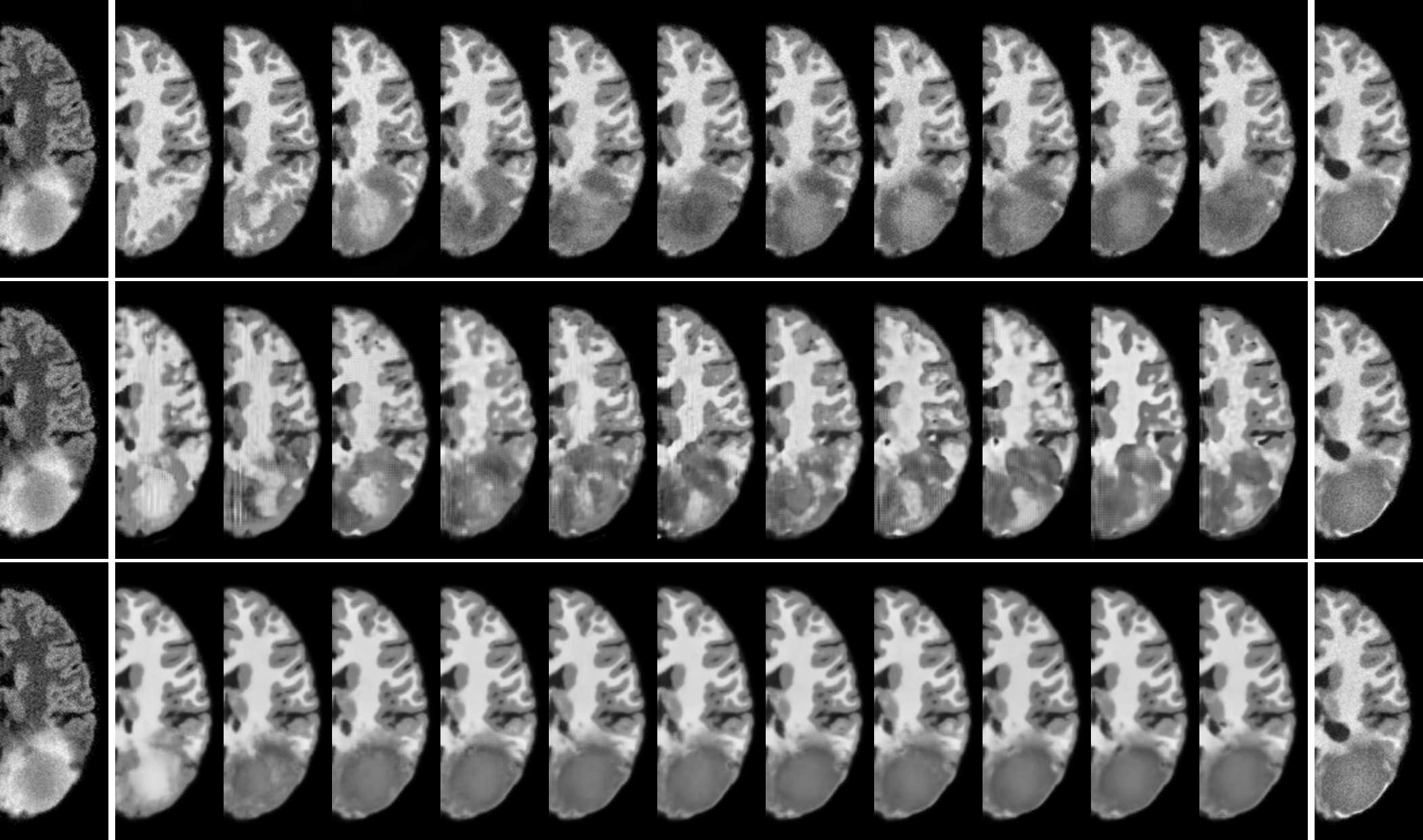}}
\fboxsep=0mm
\fboxrule=1pt
{\includegraphics[width=0.48\textwidth]{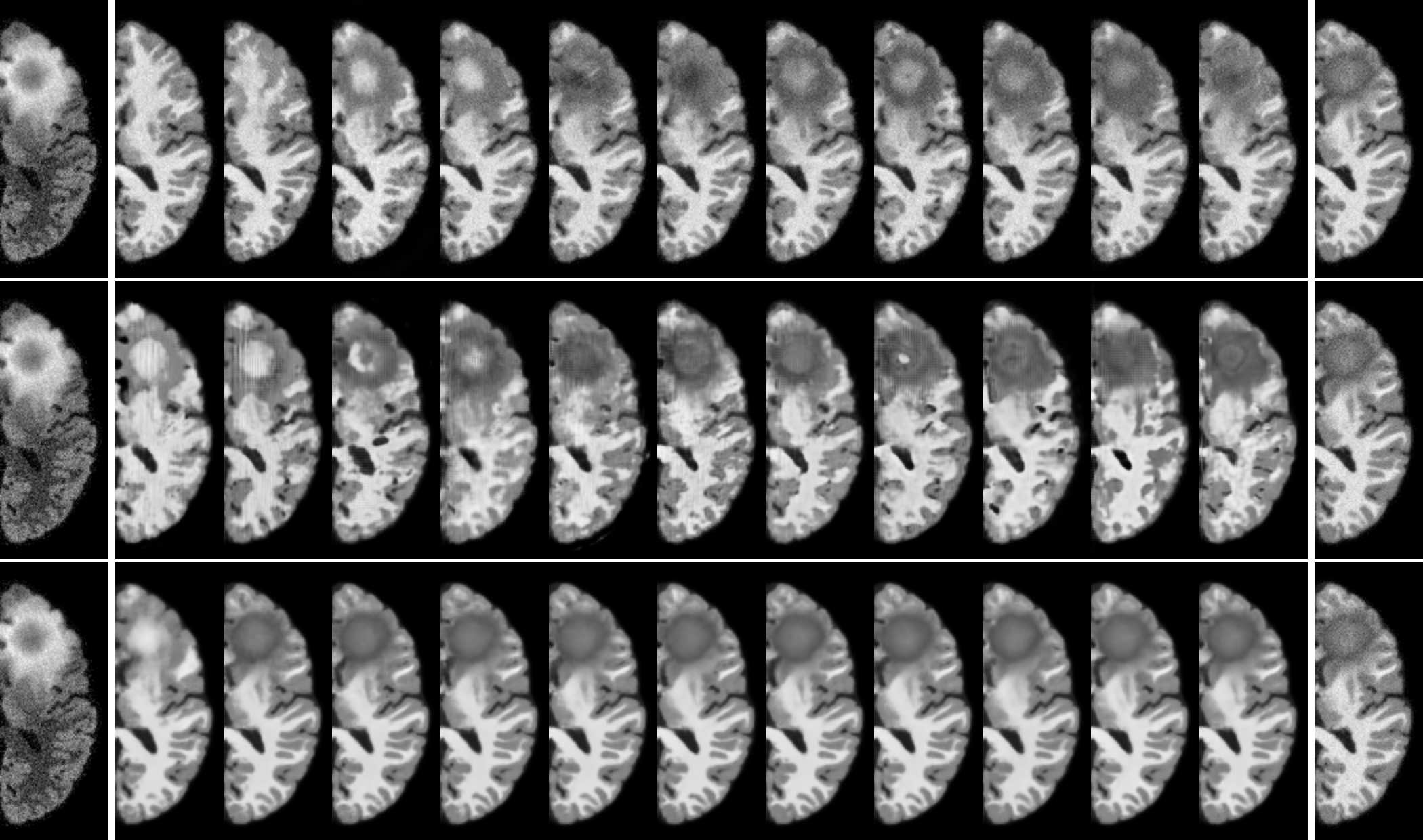}} \\
\fboxsep=0mm
\fboxrule=1pt
{\scalebox{.5}{{\hspace{15pt} Flair \hspace{4pt} 0\% \hspace{4pt} 10\% \hspace{4pt} 20\% \hspace{3.5pt} 30\% \hspace{3pt} 40\% \hspace{2.5pt} 50\% \hspace{2pt} 60\% \hspace{2pt} 70\% \hspace{2pt} 80\% \hspace{1pt} 90\% \hspace{1pt} 100\% \hspace{3pt} T1
\hspace{11pt} Flair \hspace{4pt} 0\% \hspace{4pt} 10\% \hspace{4pt} 20\% \hspace{3.5pt} 30\% \hspace{3pt} 40\% \hspace{2.5pt} 50\% \hspace{2pt} 60\% \hspace{2pt} 70\% \hspace{2pt} 80\% \hspace{1pt} 90\% \hspace{1pt} 100\% \hspace{3pt} T1}}}
\end{subfigure} 
\end{center}
\vskip-10pt
    \caption{\small Showing translation bias when the classifier detects the class wrongly. On the top row, two examples are shown where the source image is healthy and the classifier predicts tumor for all 33 translated images across all 3 losses (CycleGAN, CondGAN, L1). Despite having an imperfect translation, which leads to a tumor-like shape in the translation, see how the degree of bias is different across different loss functions. 
    On the bottom two rows, four samples are shown where the source image has a tumor and the classifier predicts healthy for all 33 translated images. 
    While classifier is doing a poor job here, there is a bias in the translated samples compared to the ground truth T1 samples.
    L1 suffers less from artifacts.
    These images show when the classifier detects the class wrongly, which can be due to having a weak classifier or imperfect translation, the bias in translation can still be observed.}
\label{fig:bias_class_diff}
\end{figure}

\end{document}